\documentclass[journal,twoside]{IEEEtran}
\usepackage{orcidlink}
\usepackage{cite}
\usepackage{hyperref} 
\hypersetup{hidelinks}
\usepackage{color}
\usepackage{amsmath}
\usepackage{multicol}
\usepackage{multirow}
\usepackage{booktabs}
\usepackage{graphicx}
\usepackage{subcaption}
\graphicspath{{figure/}}

\usepackage[ruled,linesnumbered]{algorithm2e}
\usepackage{framed} 
\usepackage{tcolorbox}
\usepackage{enumerate}

\ifCLASSINFOpdf

\else

\fi

\begin{document}

\title{Exploring Critical Testing Scenarios for Decision-Making Policies: An LLM Approach}

\author{Weichao~Xu\textsuperscript{\small\orcidlink{0009-0007-4691-1618}},
        Huaxin~Pei\textsuperscript{\small\orcidlink{0000-0003-4815-2778}},
        Jingxuan~Yang\textsuperscript{\small\orcidlink{0000-0001-9798-7347}},
        Yuchen~Shi\textsuperscript{\small\orcidlink{0009-0006-1021-8478}},
        Yi~Zhang\textsuperscript{\small\orcidlink{0000-0001-5526-866X}}, \IEEEmembership{Senior~Member, IEEE}
        and~Qianchuan~Zhao\textsuperscript{\small\orcidlink{0000-0002-7952-5621}}, \IEEEmembership{Senior~Member, IEEE}
\thanks{This work was supported in part by the National Natural Science Foundation of China under Grant 62133002. (\emph{Corresponding author: Huaxin Pei.})}
\thanks{Weichao Xu, Huaxin Pei, Jingxuan Yang, and Yuchen Shi are with the Department of Automation, Tsinghua University, Beijing 100084, China (email: xwc23@mails.tsinghua.edu.cn, phx17@tsinghua.org.cn, yangjx20@mails.tsinghua.edu.cn, shiyuche21@mails.tsinghua.edu.cn).}
\thanks{Yi Zhang is with the Department of Automation, Beijing National Research Center for Information Science and Technology (BNRist), Tsinghua University, Beijing 100084, China, and also with the Jiangsu Province Collaborative Innovation Center of Modern Urban Traffic Technologies, Nanjing 210096, China (e-mail: zhyi@mail.tsinghua.edu.cn).}
\thanks{ Qianchuan Zhao is with the Center for Intelligent and Networked Systems, Department of Automation, Tsinghua University, Beijing 100084, China (e-mail:zhaoqc@tsinghua.edu.cn).}} 

\maketitle

\begin{abstract}
  Recent advances in decision-making policies have led to significant progress in fields such as autonomous driving and robotics. However, testing these policies remains crucial with the existence of critical scenarios that may threaten their reliability. Despite ongoing research, challenges such as low testing efficiency and limited diversity persist due to the complexity of the decision-making policies and their environments. To address these challenges, this paper proposes an adaptable Large Language Model (LLM)-driven online testing framework to explore critical and diverse testing scenarios for decision-making policies. Specifically, we design a ``generate-test-feedback'' pipeline with templated prompt engineering to harness the world knowledge and reasoning abilities of LLMs. Additionally, a multi-scale scenario generation strategy is proposed to address the limitations of LLMs in making fine-grained adjustments, further enhancing testing efficiency. Finally, the proposed LLM-driven method is evaluated on five widely recognized benchmarks, and the experimental results demonstrate that our method significantly outperforms baseline methods in uncovering both critical and diverse scenarios. These findings suggest that LLM-driven methods hold significant promise for advancing the testing of decision-making policies.
\end{abstract}


\IEEEpeerreviewmaketitle

\section{Introduction}

\IEEEPARstart{R}{ecent} years the policies solving sequential decision-making problems have achieved promising results across various fields such as autonomous driving \cite{chen2020learning,toromanoff2020end,chib2023recent,10244078}, robotics \cite{kober2013reinforcement,duan2022survey,shi2024}, and Go \cite{silver2016mastering}. With advancements in artificial intelligence technology, neural network-based decision-making policies are now capable of matching or even surpassing human performance.  
Despite their remarkable effectiveness, these policies face challenges related to interpretability, robustness, and reliability, leading to critical scenarios that may trigger failures during application.
To address these challenges, it is essential to develop efficient and comprehensive testing methods for evaluating the performance of decision-making policies.

Testing decision-making policies involves uncovering a series of states that represent critical scenarios, which expose incorrect behaviors in the target policy and cause it to fail at its task. However, since these policies are predominantly developed using neural networks, their ``black-box'' nature poses challenges to understanding and predicting their behaviors. Furthermore, the continuous interactions between the policy and its environment add complexity to predicting critical scenarios. Additionally, the potentially infinite number of states and the high-dimensional state space involved can complicate testing, further hindering the search for critical scenarios.

Increasing attention has been given to developing testing methods to address the challenges outlined above\cite{10422684,feng2023dense,riedmaier2020survey,biagiola2024boundary,abdessalem2018testing,julian2020validation}. However, a universal and efficient testing framework is still lacking. Existing methods often rely heavily on complex designs and prior knowledge, which limits their adaptability to large-scale and diverse tasks. Moreover, the flexibility of these frameworks is constrained by their specific design elements, restricting their capacity for self-evolution. While simplifying designs to improve generalizability can enhance adaptability, it may compromise the effectiveness of testing methods, preventing the simultaneous achievement of both efficiency and universality. Given the diversity and ``black-box" nature of the policies under test, an efficient online testing method is required—one that can dynamically adapt to the observed performance of the target policies.

The emergence of Large Language Models (LLMs) has recently introduced innovative approaches to tackling complex problems across diverse fields \cite{sha2023languagempc,sun2024adaplanner,li2022competition}. By leveraging extensive training data, LLMs demonstrate a remarkable level of intelligence, encompassing broad world knowledge and advanced reasoning capabilities. In the context of testing decision-making policies, LLMs can utilize their inherent common sense and reasoning skills to analyze challenging scenario characteristics across various environments. This enables adaptive scenario generation with minimal human intervention. Furthermore, techniques such as prompt engineering \cite{sahoo2024systematic} allow LLMs to learn iteratively from past testing experiences, supporting efficient online testing. Their creative potential further enhances testing by generating novel and diverse scenarios, even when constrained by limited initial inputs or reference cases. By creatively modifying scenarios at multiple scales, LLMs enable cross-regional exploration and uncover previously uncharted areas.

Building on the strengths of LLMs, this paper investigates how their reasoning and learning capabilities can be harnessed to generate critical testing scenarios, enabling efficient and versatile testing of decision-making policies. However, several key challenges emerge when developing an LLM-driven approach: \emph{i)} In the absence of well-defined benchmarks, LLM-generated responses often lack depth and complexity, making it challenging to directly create intricate scenarios; \emph{ii)} Inherent limitations, such as hallucinations, pose a challenge in guiding LLMs to focus on the target environment, reason accurately, and evolve based on human input or historical experience; \emph{iii)} The probabilistic nature of LLM text generation results in responses biased toward common patterns, leading to outcomes that may lack precision or nuance. For instance, prompting an LLM to generate a specific value like 51.7, as opposed to a rounded value like 50, within a defined range (e.g., [0,100]), remains a notable challenge.

\begin{figure}[t]
    \centering
    \centerline{\includegraphics[width=0.9\linewidth]{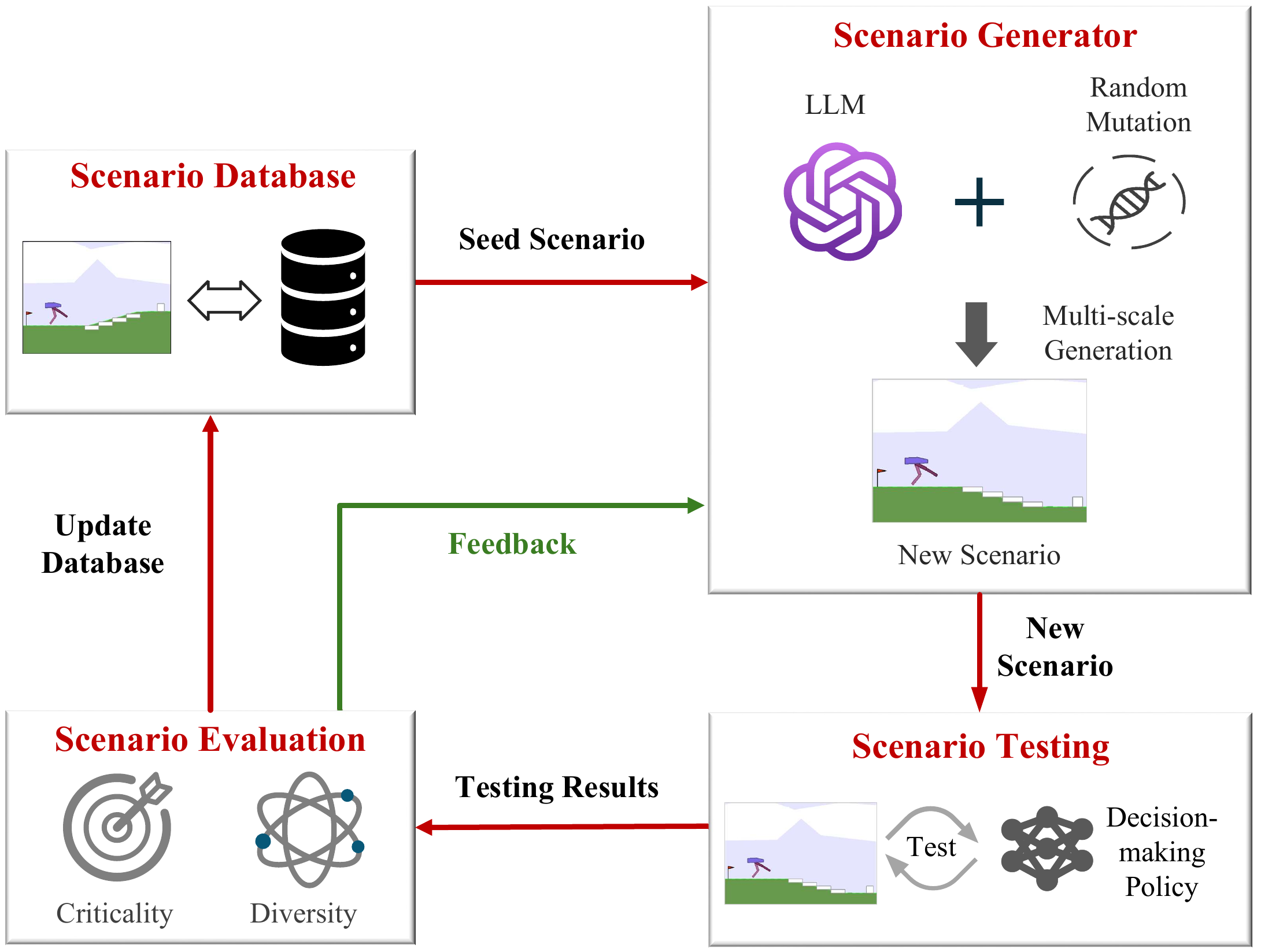}}
    \caption{The overview of LLM-driven online testing framework.}
    \label{figure:simplied_framework}
\end{figure}

To overcome the above challenges, this paper introduces LLMTester, an innovative LLM-driven testing framework designed to evaluate decision-making policies. The proposed framework illustrated in Fig. \ref{figure:simplied_framework} integrates three key components: a scenario database, a scenario generator powered by LLMs, and a generalized evaluation and feedback mechanism. Together, these components form a dynamic ``generate-test-feedback" pipeline, enabling flexible and adaptive testing guided by diverse inputs. The scenario database plays a critical role by providing and maintaining seed scenarios that act as benchmarks for the LLM-driven generator. By anchoring the generator to these benchmarks, the framework prevents random or trivial outputs from LLMs, ensuring the consistent creation of challenging scenarios.

To mitigate hallucinations and enable automation, a carefully designed prompt template is developed in this paper. This template not only minimizes human involvement but also incorporates techniques such as Chain-of-Thought (COT) \cite{wei2022chain}, which effectively guides the LLMs toward deeper reasoning and analysis. Following prompt engineering, the LLM-based scenario generator quickly comprehends the target environment and leverages feedback from historical experience and expert knowledge to generate critical scenarios.

Moreover, it has been observed that LLMs often exhibit insufficient performance in terms of the level of refinement in scenario generation, limiting their ability to make fine-grained adjustments. To address this issue, we propose a multi-scale generation strategy. This strategy adaptively determines whether to use the LLM-based generator or apply random perturbations to the seed scenario, based on an analysis of whether the given scenario is of high potential. As a result, the strategy effectively identifies critical scenarios while reducing resource consumption, thus addressing the inherent challenges LLMs face in making fine adjustments.

The proposed LLM-driven method is evaluated on five widely recognized benchmarks, and the results indicate that our proposed LLM-driven method significantly outperforms the baseline in terms of the discovery rate of critical scenarios, while also ensuring the diversity of these failure scenarios. Further ablation and comparative experiments highlight the effectiveness and robustness of our approach. 

To sum up, our contributions are as follows: \emph{i)} We present a general LLM-driven online testing framework with a ``generate-test-feedback'' pipeline, enabling efficient testing of decision-making policies. \emph{ii)} A user-friendly and effective prompt template for LLM-driven testing scenario generation is developed to enable efficient and rapid testing of new policies, based on an analysis of the characteristics of decision-making tasks. \emph{iii)} To address the limitations of LLMs in making fine-grained adjustments, a multi-scale generation strategy with adaptive potential analysis is proposed, which further boosts testing efficiency while reducing resource consumption. \emph{iv)} Our experimental results, derived from testing five policies across four distinct environments, demonstrate the effectiveness and general applicability of the proposed method.

The rest of this paper is organized as follows. Section \ref{sec:relatedworks} reviews the related work. In Section \ref{sec:methodology}, we introduces the details of our method, including the online testing framework, LLM-based scenario generator, and multi-scale generation strategy. Section \ref{sec:exp} describes the experimental setup and evaluation results, while Section \ref{sec:discussion} presents the discussions and threats to validity. Finally, Section \ref{sec:conclusion} concludes the paper.

\section{Related Work} \label{sec:relatedworks}
\subsection{Testing of Decision-making Policies}

\subsubsection{Input Modification} Testing decision-making policies involves generating test cases for the policies to address, with the primary objective of identifying cases that lead to failure. The most straightforward approach is to directly modify the inputs to the decision-making policies.
Various testing methods \cite{pei2017deepxplore,guo2018dlfuzz,ma2018deepmutation,zolfagharian2023search,tian2018deeptest,zhang2018deeproad} focus on designing algorithms for finding special test inputs to evaluate the security and robustness of DNN-based policies. DeepXplore \cite{pei2017deepxplore} introduces the concept of neuron coverage, and generates test inputs through gradient ascent to uncover test cases that achieve high neuron coverage and elicit differing behaviors across various DNNs. However, it is limited to white-box settings, which restricts its scope of application. DLFuzz \cite{guo2018dlfuzz} employs a fuzz testing framework, where it mutates inputs by optimizing a specially designed objective function aimed at reaching higher neuron coverage and exposing more exceptional behaviors. STARLA\cite{zolfagharian2023search} introduces a data-box testing approach and uses a genetic algorithm to modify the episodes of reinforcement learning (RL) agents, aiming to detect more faults. In autonomous driving tasks, DeepTest\cite{tian2018deeptest} uses neuron coverage as a guiding metric to modify input images for decision-making policies, thereby generating new test cases. Furthermore, DeepRoad\cite{zhang2018deeproad} leverages a generative adversarial network (GAN) trained on large data pairs, enabling the GAN to directly learn and generate transformed images. 

\subsubsection{Scenario Generation} In some constrained environment, the casually generated inputs of neural networks may not correspond to real-world scenarios, making such inputs practically irrelevant. Additionally, since internal policies are often highly encapsulated, many approaches\cite{duan2020test,feng2021intelligent,koren2018adaptive,du2019finding,delecki2022we,lu2022learning,zhong2022neural,huai2023sceno,ma2024diversity,zhao2016accelerated,feng2020testing,yan2023learning} test decision-making policies through scenario generation. For instance, \cite{koren2018adaptive,du2019finding,delecki2022we,lu2022learning} employ methods based on RL and Monte Carlo tree search to perform adaptive stress testing on autonomous driving, thereby generating challenging test scenarios. To identify unique traffic violations, Autofuzz \cite{zhong2022neural} uses a neural network for seed selection and mutation, and implements a grammar-based fuzzing framework. Meanwhile, scenoRITA \cite{huai2023sceno} introduces evolutionary algorithms to search for critical scenarios.  For testing competitive game agents, AdvTest\cite{ma2024diversity} designs and adds constraints to guide adversarial agent training, with the goal of exposing a more diverse range of failure scenarios. However, the parameter setting and training of these methods require additional testing costs. In parallel, data-driven scenario generation methods have been continuously proposed based on extensive training data. \cite{zhao2016accelerated,feng2020testing} construct new scenario datasets by analyzing the entire scenario data and sampling safety-critical scenarios, while NeuralNDE \cite{yan2023learning} utilizes a Transformer-based model to learn from real-world data and efficiently generate synthetic data with distributions that closely similar to real-world distributions.

However, the methods mentioned above either place high demands on the tested policies, such as requiring white-box access or large datasets, or are tailored to specific tasks. Therefore, a general testing framework for decision-making policies is urgently needed. MDPFuzz \cite{pang2022mdpfuzz} proposes the first general-purpose fuzz testing framework for models solving MDPs. Building upon MDPFuzz, SeqDivFuzz \cite{wang2023fuzzing} enhances testing efficiency by terminating non-diverse sequences early, based on sequence diversity inference. GMT \cite{li2023generative} trains a generative diffusion model to generate diverse, failure-triggering test cases for decision-making policies and continuously fine-tunes the generative model during testing. For multi-agent systems, MASTest \cite{ma2024enhancing} calculates the criticality of states (considering both exploitation and exploration) and perturbs the action at critical states. However, the pursuit of generality in these methods comes at a cost: they do not account for the specific characteristics of the environment under test, which limits the efficiency of the testing process. Large Language Models (LLMs), with their powerful comprehension abilities and extensive knowledge base, are capable of generating targeted test scenarios with minimal human input. Leveraging LLMs, our proposed approach can achieve both efficient and universal scenario generation.

\begin{figure*}[htbp]
  \centering
  \includegraphics[width=\textwidth]{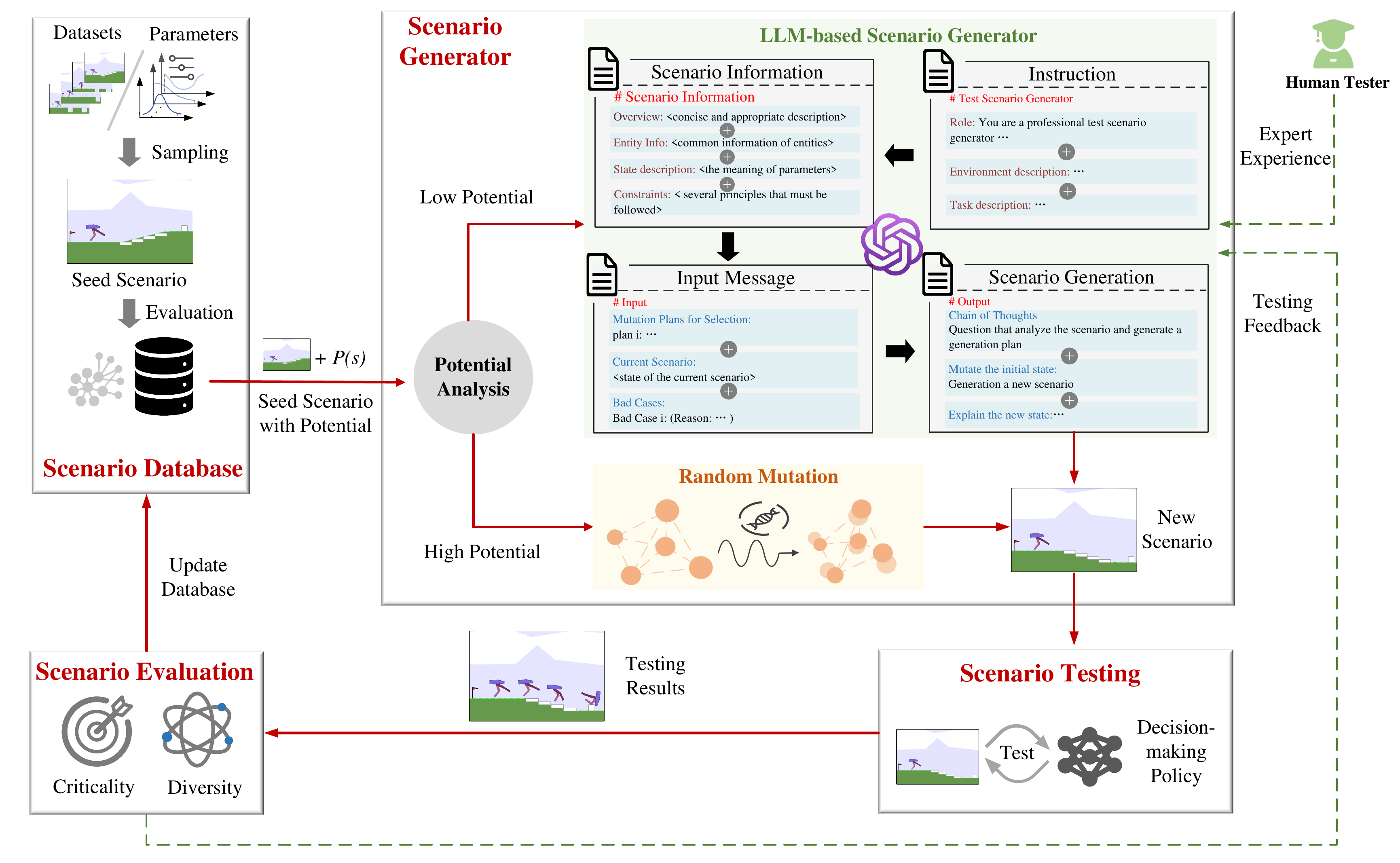}
  \caption{The workflow of our LLM-driven online testing framework. \textit{Scenario Database} provides the seed scenario as a reference for generating new scenarios.  \textit{Scenario Generator}, through prompt engineering and a multi-scale generation strategy, gives full play to the intelligence of LLM to generate critical scenarios efficiently. \textit{Scenario Testing} tests the policy in the given scenarios. \textit{Scenario Evaluation} assesses the criticality and diversity based on the testing results, providing feedback to the LLM for self-improvement.}
  \label{fig:method}
\end{figure*}

\subsection{Testing with LLMs}
As LLMs have been applied across various fields,  \cite{wang2024software,yu2023llm,kang2023large,deng2023large,liu2024testing} is seeking to utilize the capabilities of LLMs to produce test cases for the purpose of testing a specific algorithm or software. For example, LIBRO \cite{kang2023large} proposes using LLMs to reproduce a given bug by querying the model to generate test methods that align with the bug report. TitanFuzz \cite{deng2023large} applies fuzzy testing to deep learning libraries, using LLMs' code generation and completion abilities to mutate seeds. As a result, it successfully identifies previously unknown bugs in TensorFlow/PyTorch.  InputBlaster \cite{liu2024testing} first utilizes LLMs to generate valid text, then applies mutation rules to generate unusual inputs that trigger application crashes. 

Similarly, LLMs have been widely developed for scenario generation in areas such as autonomous driving. ChatScene \cite{zhang2024chatscene} uses an LLM to generate descriptions of safety-critical scenarios and retrieves corresponding Scenic code snippets from a pre-constructed database, though the design and acquisition of Scenic code remain relatively challenging, limiting its applicability and generalization to other tasks. ChatSim \cite{wei2024editable} engages multiple LLM agents in collaborative efforts to understand user commands and generate corresponding photo-realistic scene videos, but it is designed for generating videos based on user requests, rather than for producing critical scenarios. SeGPT \cite{li2024chatgpt} leverages real-world scenarios, allowing an LLM to modify them and generate synthetic datasets for trajectory prediction testing, with a primary focus on modifying vehicle trajectories for datasets tailored to evaluating and training trajectory prediction algorithms. LLMScenario \cite{chang2024llmscenario} extracts naturalistic risky trajectories from a database and guides the LLM to produce different challenging trajectories, including the ego vehicle’s trajectory, but it is not a specialized testing framework, as it does not address challenges posed by the black-box nature of the target policies. OmniTester \cite{lu2024multimodal} takes a further step by employing multimodal LLMs to generate road networks and vehicle configurations, understanding user testing requirements and generating controllable, realistic, and challenging scenarios within a carefully designed framework. It generates test scenarios by constructing road layouts and positioning vehicles using tools like SUMO, which imposes strict constraints on scenario formats and environments. 
The methods described above lack an effective online optimization framework during scenario generation, limiting their ability to identify flaws in policies—--especially in tasks where predicting the agent’s behavior is more complex than in autonomous vehicle applications. Consequently, the scenarios generated by these methods may not always be sufficiently challenging. 

To summarize, these methods are primarily designed for specific domains, particularly in autonomous driving and software testing, and do not address generalized scenarios. This limitation results in a lack of testing capability for more universal systems. In contrast, our approach is aimed at general decision-making policies, outlining a scenario generation template and guiding LLM-driven scenario generation through an online testing framework.

\section{Methodology} \label{sec:methodology}
We propose a general LLM-driven online testing approach to efficiently test decision-making policies by exploring critical scenarios. 
In this section, we first introduce the overall testing framework with four modules in Section \ref{subsec:framework}. In Section \ref{subsec:scenario generator}, we focus on the LLM-based scenario generator within this framework. Finally, to address the inherent challenges faced by LLMs and further enhance the testing efficiency, a multi-scale generation strategy is proposed in Section \ref{subsec:multi-scale}.

\subsection{The Testing Framework} \label{subsec:framework}
Fig. \ref{fig:method} shows the overall procedure of our LLM-driven testing framework. The whole framework can be divided into four modules: \textit{Scenario Database}, \textit{Scenario Generator}, \textit{Scenario Testing} and \textit{Scenario Evaluation}. Each of these modules is designed with the flexibility to adapt to different policies and environments.

\subsubsection{Scenario Database} 
The process of this framework begins with the establishment of a scenario database, which stores key parameters representing each scenario (such as the initial state), along with associated testing information. Each scenario in the database serves as a seed for further modification by the scenario generator. During testing, seed scenarios are continuously selected and provided to the generator to create new scenarios, while the accompanying test information helps enhance the generator’s capabilities. 
In this way, the generation and maintenance of the scenario database can be refined through various guidance techniques to meet different testing objectives. For instance, the database can be updated to include only scenarios with higher criticality and greater diversity. This update process is driven by the performance of the policy under test, ensuring that the seed scenarios provide dynamic and relevant information for the policy.

Without loss of generality, the scenario database can be constructed through random sampling after parameterizing the elements within the environment, or by sampling from pre-existing datasets. For example, in autonomous driving, testing scenarios can be initialized by randomly sampling the vehicle's location and yaw, or by extracting data from datasets such as KITTI\cite{geiger2012we}.

\subsubsection{Scenario Generator} 
The scenario generator, upon receiving seed scenarios sampled from the database, generates new testing scenarios based on the seeds. Simultaneously, it incorporates testing feedback from the scenario evaluation module to refine the generation process.
The core component of this module is an LLM-based scenario generator. The LLM-based generator transforms seed scenarios, testing feedback, and expert knowledge into prompts using a standardized template, thereby guiding the LLM to generate new and more challenging scenarios.
In this paper, the scenario generator is also capable of creating scenarios through random mutation. Specifically, a multi-scale generation strategy is implemented to combine the advantages of both the LLM-based generator and random mutation, enhancing the efficiency of exploring critical testing scenarios.
The details of the scenario generator can be found in Section \ref{subsec:scenario generator} and \ref{subsec:multi-scale}.

\subsubsection{Scenario Testing}
The new scenarios generated by the scenario generator will be used to configure the environment and conduct a test for the target policy. Subsequently, the actual performance of the policy under test will be evaluated and utilized for online testing.

\subsubsection{Scenario Evaluation} 
The scenario evaluation module assesses both seed scenarios and newly generated ones across two primary dimensions: criticality and diversity. These evaluations guide three key processes: selecting seed scenarios, updating the scenario database, and generating new scenarios. The measurement of criticality and diversity utilizes specific metrics designed to align with the performance of the policies being tested. Based on the evaluation results, the module determines the importance of each scenario and decides whether it should be sampled or included in the database. Additionally, this testing data can be fed back to the scenario generator to support online testing.

The evaluation module can be designed flexibly based on the target environment, as long as the evaluation scores are correlated with the importance of the scenarios. Here, we recommend the evaluation method proposed in MDPFuzz \cite{pang2022mdpfuzz}. When selecting seed scenarios, sensitivity is computed as the sampling weight by randomly mutating the initial state of the seed scenario, and is calculated as:
\begin{equation}
  \rho = \frac{|r_{\mathrm{seed}} - r_{\delta}|}{||\Delta||_2},
\end{equation}
where $r_{\mathrm{seed}}$ and $r_{\delta}$ are the cumulative rewards obtained by the decision-making policy under seed scenario and mutated scenario, respectively. $\Delta$ represents a random permutation. 
A seed scenario with higher sensitivity is more likely to exhibit significant changes in response to small perturbations, thereby increasing the likelihood of failure when new scenarios are generated from it. When updating the database, freshness and cumulative reward, as defined in MDPFuzz, are used to determine whether a new scenario should be added. Furthermore, a specialized evaluation mechanism for online testing will be employed in scenario generation.

In the following sections, we provide a detailed introduction to the scenario generator, including the LLM-based scenario generator (Section \ref{subsec:scenario generator}) and a multi-scale generation strategy (Section \ref{subsec:multi-scale}).

\begin{figure}[htbp]
\centering
\centerline{\includegraphics[scale=0.45]{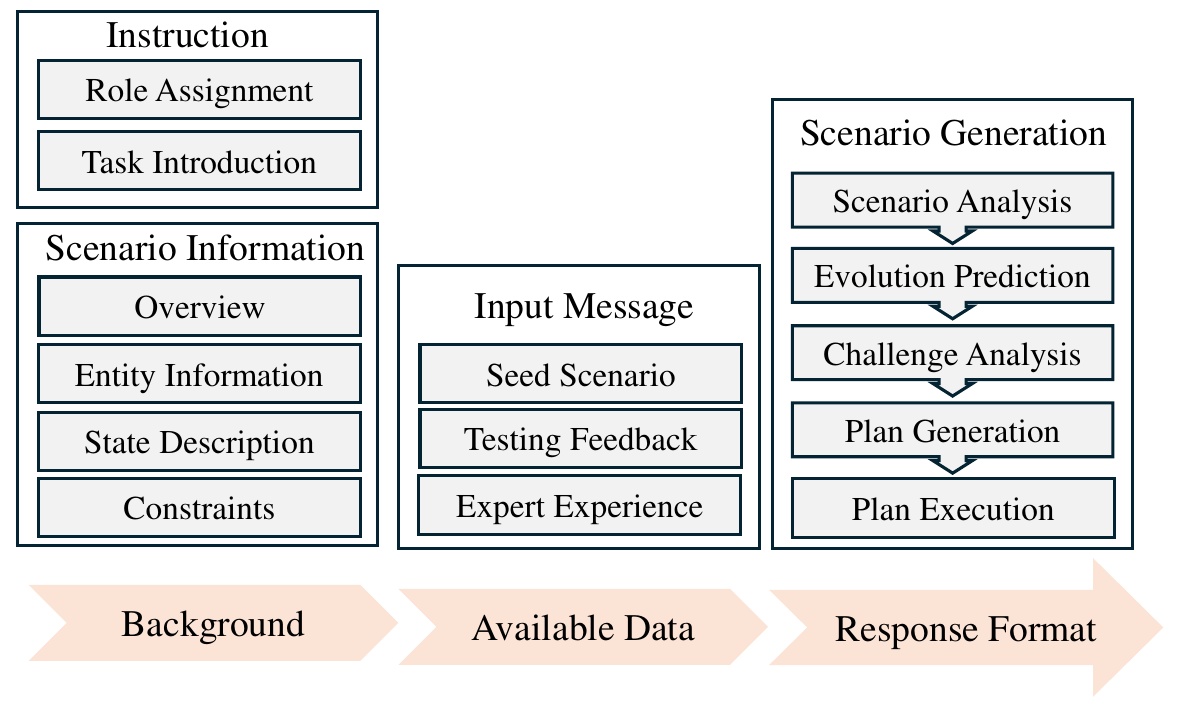}}
\caption{Key elements for designing a prompt.}
\label{figure:elements}
\end{figure}

\subsection{LLM-based Scenario Generator} \label{subsec:scenario generator}
The LLM-based scenario generator operates using environment-specific prompts and integrates expert knowledge along with feedback from prior testing. It modifies seed scenarios to create new ones with a higher likelihood of inducing failures. The process is outlined as follows:
\begin{equation} \label{Equ snew}
    s_{\mathrm{new}} = \mathcal{G}_{\mathrm{env}}(s_{\mathrm{seed}},F,E),
\end{equation}
where $ \mathcal{G}_{\mathrm{env}}$ denotes that the LLM-based scenario generator adapts to specific environments. Environment-specific prompts assist the LLM in understanding its tasks and target environment, guiding the generation of new scenarios. During testing, the seed scenario $s_{\mathrm{seed}}$, expert experience $E$, and historical testing feedback $F$ are automatically incorporated into the prompts. $s_{\mathrm{seed}}$ serves as a reference for scenario generation, while $E$ enhances the LLM's reasoning capabilities with human expertise. Historical feedback $F$ is crucial, as it enables the LLM to generate scenarios through online optimization, allowing it to identify more critical scenarios relevant to the policies under test.

To minimize the effort required for prompt engineering, we analyze the characteristics of decision-making tasks and propose a workflow that enables efficient prompting of an LLM, as shown in Fig. \ref{figure:elements}. For effective online testing, the prompt must include environment-specific background knowledge, available data during the testing process, and the response format that the LLM needs to follow.
Furthermore, we refine and identify several key elements to include in the prompt, categorizing them into four aspects of tokens: instruction, scenario information, input message and scenario generation.

\subsubsection{Instruction} 
To help an LLM quickly focus on a specific task, it is common practice to assign a role to the LLM at the beginning of the prompt and clearly define its task. Consequently, the instruction tokens consist of two key elements: \textit{Role Assignment} and \textit{Task Introduction}. These tokens enable the LLM to gain a clear understanding of both its role and task, allowing it to generate more targeted responses throughout the Q\&A process.

\subsubsection{Scenario Information} 
Detailed scenario information should be provided to the LLM to help it better understand the key aspects of the target environment.  When testing general decision-making policies, scenario information includes the following components:  \textit{Overview}, \textit{Entity Information}, \textit{State Description}, and \textit{Constraints}. \textit{i) Overview} offers a concise and accurate description of the target environment, outlining the agent(s)' tasks, the definitions of crash or failure, and other essential details. \textit{ii) Entity Information} covers the common characteristics (e.g., number, type, physical properties, and interactions) of the entities within the environment. These entities include not only the agent(s) controlled by the target policies but also any elements that may interact with the agent(s), such as obstacles. \textit{iii) State Description} explains the meaning of the scenario's parameters provided in the subsequent input message. \textit{iv) Constraints} lists the principles that must be followed during scenario generation. These constraint tokens help standardize the output from the LLM, ensuring that the newly generated scenarios are valid and solvable. Scenario information tokens provide general details about the target environment, which remain constant throughout the testing process. As a result, the tester only needs to design the scenario information once.

\begin{figure}[htbp]
\centering
\centerline{\includegraphics[scale=0.8]{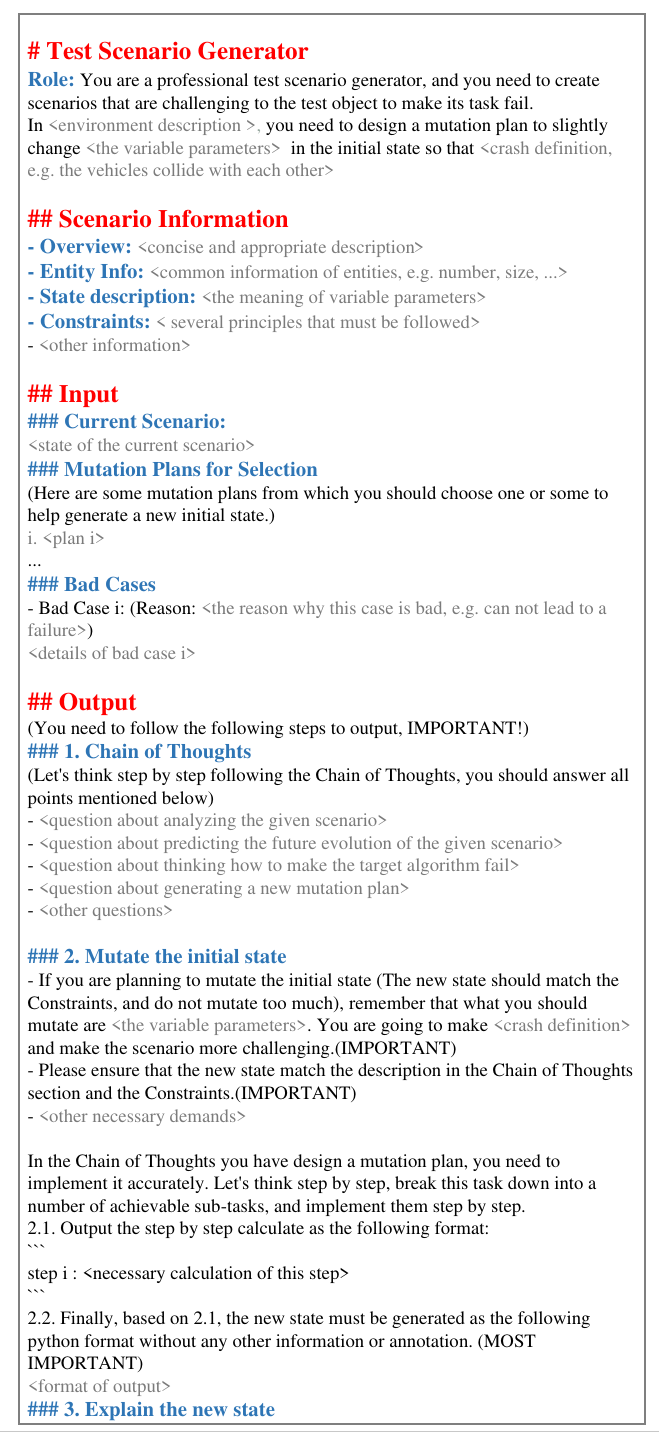}}
\caption{The Prompt Template. \textit{Instruction} (Role) provides the overview of LLM's task and the target environment. \textit{Scenario Information} (\#\#Scenario Information) outlines the common and unchanging information of the target environment. \textit{Input Message} (\#\#Input) includes the variable information during testing, such as the seed scenario, feedback, and expert experience. \textit{Scenario Generation} (\#\#Output) guides the LLM to generate and output a new scenario in a specified format.}
\label{figure:prompt}
\end{figure}

\subsubsection{Input Message} 
The input message tokens contain available data that vary as the scenario changes, corresponding to $s_{\mathrm{seed}}$, $F$ and $E$ in Eq. (\ref{Equ snew}). These tokens serve as input interfaces for \textit{Seed Scenario}, \textit{Testing Feedback}, and \textit{Expert Experience}.

\subsubsection{Scenario Generation}
Based on the background information and available data, the LLM is tasked with thoroughly understanding the scenario and ultimately generating one that meets the testing requirements. To achieve this, the prompt for scenario generation can guide the LLM through the following workflow: \textit{i) Scenario Analysis}: describing and analyzing the initial state of the given seed scenario, including the agents' states and the relationships between entities; \textit{ii) Evolution Prediction}: predicting the future evolution of the scenario, such as the trajectories and interactions between entities; \textit{iii) Challenge Analysis}: identifying how to make the decision-making task fail and providing the corresponding idea; \textit{iv) Plan Generation}: generating a plan to modify the seed scenario; \textit{v) Plan Execution}: executing the plan correctly and outputting a new scenario in the same format as the input.

By integrating the key elements outlined in Fig. \ref{figure:elements}, a prompt can be efficiently designed without introducing excessive workload. Various prompting techniques and styles can be combined to create the prompt, and Fig. \ref{figure:prompt} presents a prompt template used in this paper, which is described in detail below.

We first assign the LLM the role of a ``test scenario generator'' and specify that its task is to mutate the state of a given scenario in such a way that the target policy fails. Next, we introduce the basic information about the environment, the meaning of the initial state, and the constraints governing scenario generation. For the input message, the initial state of the seed scenario is transformed into the format described in the scenario information. 

Bad cases encountered during testing are considered as $F$ and incorporated into the prompt. In this paper, three criteria are used to classify a new scenario as a bad case: \emph{i) Insufficient Challenge}: This type of bad case is recorded when there is a significant increase in reward after generation, i.e., $r_{\mathrm{new}}-r_{\mathrm{seed}}>\mathcal{T}_r$, where $\mathcal{T}_r$ is a fixed threshold. This indicates that the LLM is not mutating the given scenario in a way that increases its difficulty. \emph{ii) Invalidity}: This type of bad case is recorded when the new scenario breaks the constraints or is unsolvable even by an optimal policy. \emph{iii) Excessive Modification}: This type of bad case is recorded when the difference between the new scenario and the seed scenario exceeds a specified threshold $\mathcal{T}_s$, i.e., $||s_{\mathrm{new}}-s_{\mathrm{seed}}||>\mathcal{T}_s$. This criterion is applied when the tester expects the LLM to generate scenarios that are similar to the seed scenario.

The expert experience $E$ in this template follows the format of mutation plans. With this format, the LLM can easily reference expert experience and generate its own plan based on the expert's plan. The mutation plan can be very simple, requiring minimal effort from the tester (e.g., slightly mutate the seed to make the scenario more challenging). Alternatively, it can be more detailed, based on prior knowledge and experience.

When guiding the LLM to generate outputs, we employ COT \cite{wei2022chain} to gradually refine its understanding of the scenario into a feasible mutation plan. COT is an important prompting technique that enables LLMs to reason coherently through a series of step-by-step processes. The LLM is then instructed to break the scenario generation task down into a series of achievable sub-tasks, facilitating the execution of the proposed mutation plan. The final output is structured to match the format of the input state, ensuring easy extraction during post-processing. Finally, we prompt the LLM to explain the scenario it generates, encouraging it to reconsider its approach and produce more reliable and reasonable results. This explanation also helps testers understand the rationale behind the LLM’s generated scenario.

\begin{figure}[t]
\centering
\centerline{\includegraphics[width = \linewidth]{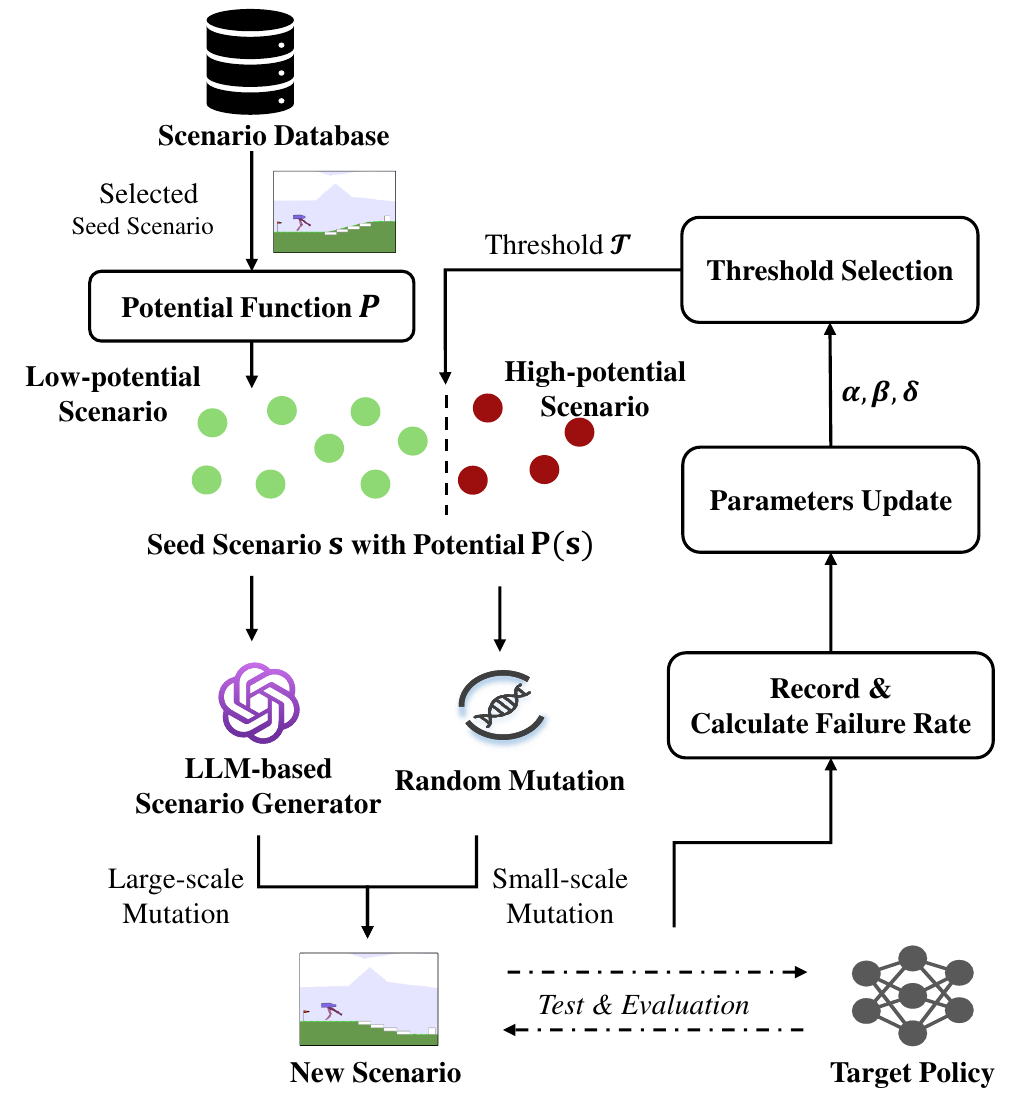}}
\caption{The illustration of multi-scale generation strategy.}
\label{fig:multi-scale}
\end{figure}

\subsection{Multi-scale Generation Strategy} \label{subsec:multi-scale}
Despite their strong performance, LLMs are known to struggle with fine-grained or complex tasks, such as those that require high computational precision. In the context of critical scenario generation, this limitation results in poor performance when fine adjustments are made to seed scenarios, undermining their overall effectiveness. To address these challenges, we propose a novel multi-scale scenario generation strategy which enhances efficiency while conserving valuable testing resources. The workflow of this strategy is illustrated in Fig. \ref{fig:multi-scale}.

\subsubsection{Motivation} 
We note that LLM-driven scenario generation establishes a direct link between the generator and the target environment, making it highly targeted and efficient. However, despite their computational capabilities, LLMs are not particularly adept at performing precise mathematical calculations. Additionally, LLMs inherently exhibit a discretization characteristic in mutation amplitude, which limits their capacity for local exploration. This limitation arises because LLMs tend to output numbers that are contextually common or typical. For instance, when asked to modify the number 1.5 within a range of 0 to 10, an LLM is more likely to produce values such as 1 or 2, rather than more granular adjustments like 1.7. As a result, LLMs show limited sensitivity to small distinctions in numerical values and struggle to capture subtle differences between scenarios. Consequently, such generators are not well-suited for effectively utilizing near-failure scenarios, which are only one step away from critical situations. In contrast, a small and random mutation can provide fine-grained variation to the seed scenario, resulting in better local exploration ability. 
Building on the background above, we propose a multi-scale generation strategy that enables large-scale mutations using the LLM in general scenarios, while applying small-scale random mutations in near-failure scenarios. 

\subsubsection{Scenario Potential Analysis}
We determine the type of mutation applied to a scenario by analyzing its potential. Specially, each scenario is characterized by its distance from critical scenarios, reflecting its potential to escalate into a critical scenario following random perturbations. A higher potential indicates a greater likelihood of transitioning into a critical scenario with minor perturbations. We define \emph{High-potential Scenarios} as those that are near-failure and more likely to evolve into critical scenarios, potentially resulting in a crash. To quantify the potential, we introduce a potential prediction function $P$, where the potential of a given scenario is calculated as $P(s)$. By setting an appropriate threshold $\mathcal{T}$, scenarios with potential values exceeding this threshold are classified as high-potential scenarios. The set of high-potential scenarios is then represented as 
\begin{equation}
    \mathcal{H} = \{s | P(s) \geq \mathcal{T}\}.
\end{equation}
On the contrary, the set of \emph{Low-potential Scenarios} is as
\begin{equation}
    \mathcal{L} = \overline{\mathcal{H}} =  \{s | P(s)<\mathcal{T}\}.
\end{equation}
We reuse the cumulative rewards $r$ discussed in \textit{Scenario Evaluation} module in Section \ref{subsec:framework} as the measure for high-potential scenarios, as these rewards are strongly correlated with the quality of task completion by the policy being tested within each scenario. Here, $P(s) = -r_s$, and scenarios with rewards below a given threshold are classified as high-potential scenarios.

\begin{algorithm}[t]
  \caption{Scenario Generation with Multi-scale generation strategy}
  \label{algo: MSG}
  \KwIn{current seed scenario $s_{\mathrm{seed}}$ and its potential $P_s$; $P_D$ which contains potential of each scenario of the scenario database; scenario generator $G$ with parameters $\alpha,\beta, \delta$; testing feedback $F$ and expert experience $E$ }
  \KwOut{new scenario after generation $s_{\mathrm{new}}$}
  \SetKwFunction{Update}{Update\_parameters}
  \SetKwFunction{Percentile}{Percentile}
  \SetKwFunction{Failure}{Failure}
  \SetKwProg{Fn}{function}{:}{}
  \let\oldnl\nl
  \newcommand{\nonl}{\renewcommand{\nl}{\let\nl\oldnl}}

  \setcounter{AlgoLine}{0}
  \LinesNumbered
  \eIf{$P_s < \Percentile(P_D, 1-G.\alpha)$}
    {$s_{\mathrm{new}} \leftarrow G$.LLM\_scenario\_generation($s_{\mathrm{seed}}$, $F$, $E$)\; 
    \tcp{Apply LLM-based generator to low-potential scenarios }}
    {$s_{\mathrm{new}} \leftarrow G$.random\_mutation($s_{\mathrm{seed}}$)\;
    \tcp{Apply random mutation to high-potential scenarios }}
  \If{\Failure($s_{\mathrm{new}}$)}
    {\Update{$G$}\;}
  \KwRet{$s_{\mathrm{new}}$}

  \BlankLine
  \BlankLine
  \setcounter{AlgoLine}{0}
  \nonl \Fn{\Update{$G$}}{
    $\alpha, \beta, \delta \leftarrow G.\alpha, G.\beta, G.\delta$\;
    $rate \leftarrow Calculate\_failure\_rate()$\; 
    \uIf{$rate < (1 - \delta) * G.last\_rate$}
    { \tcp{decay $\alpha$ when failure rate decreases}
    $\alpha \leftarrow \alpha * \beta$\;
     $G.last\_rate \leftarrow rate$\;}
    \ElseIf{$rate > (1 + \delta) * G.last\_rate$}
    {\tcp{raise $\alpha$ when failure rate increases}
    $\alpha \leftarrow \alpha / \beta$\;
    $G.last\_rate \leftarrow rate$\;}
    $ G.\alpha, G.\beta, G.\delta \leftarrow \alpha, \beta, \delta$\;
    \KwRet{}
  }

\end{algorithm}

\subsubsection{Generation with Potential Analysis}
High-potential scenarios are near-failure, but the LLM-based scenario generator has limitations in fine-tuning these scenarios effectively, which prevents it from fully exploiting their potential. Therefore, we apply the LLM-based scenario generator exclusively to low-potential scenarios for large-scale, environment-specific searches. In contrast, for high-potential scenarios, we employ random mutation to conduct small-scale, localized explorations. The process outlined above is described as: 
\begin{equation}
    s_{\mathrm{new}} =\left\{
    \begin{aligned}
    & \mathcal{G}_{\mathrm{env}}(s_{\mathrm{seed}},F,E), & s_{\mathrm{seed}} \in \overline{\mathcal{H}}, \\
    & s_{\mathrm{seed}} + U(-a,a) , & s_{\mathrm{seed}} \in \mathcal{H},
    \end{aligned}
    \right.
\end{equation}
where perturbations are chosen from a uniform distribution with a maximum amplitude of $a$.

\subsubsection{Adaptive Threshold Selection}
An appropriate threshold helps identify more critical scenarios while reducing the number of LLM API calls. However, due to the limited understanding of the target environment, directly determining an appropriate threshold for identifying high-potential scenarios is not feasible. Therefore, we normalize using the scenario database, assuming that high-potential scenarios constitute a certain proportion of the database. To accommodate the continuously updated nature of the scenario database, we adaptively select this proportion during the testing process and determine the threshold based on it.

Algorithm \ref{algo: MSG} shows the details of multi-scale generation with adaptive threshold selection. 
The algorithm initializes with a threshold $\alpha$, a decay factor $\beta \in (0,1)$, and a error factor $\delta$. $\alpha$ represents a percentile, where scenarios with potential in the top $\alpha\%$ in the scenario database are considered as high-potential scenarios. The core idea of the adaptive algorithm is to calculate the failure rate and update $\alpha$ after discovering new failure cases. 
Let the new failure rate be denoted by $f'$ and the failure rate calculated the last time $\alpha$ changed be denoted by $f$. 
If $f'$ decreases compared to $f$, this indicates a decline in the quality of the scenario database (i.e., there are fewer high-potential scenarios). In this case, $\alpha$ is decayed by a fixed proportion $\beta$. Conversely, if $f'$ increases, $\alpha$ is raised by dividing it by $\beta$. To prevent fluctuations in the data, we introduce a error factor $\delta$ that tolerates a certain range of variation in the failure rate. The updating process is decribed as:
\begin{equation}
    \alpha' =\left\{
    \begin{aligned}
    & \alpha \beta, & f' < (1-\delta)f, \\
    & \frac{\alpha}{\beta} , & f' > (1+\delta)f, \\
    & \alpha , & otherwise.
    \end{aligned}
    \right.
\end{equation}
 Based on experience, $\alpha$ should be sufficiently low but also encompass the optimal threshold, such as setting it to 20. $\beta$ controls the magnitude of decay and can be set to a moderate value, such as 0.5.

With the multi-scale generation strategy, we can leverage the strengths of both LLM-based generation and random mutation. It is important to note that this strategy serves as an optional enhancement to the LLM-based scenario generator, with its effectiveness depending on the quality of the scenario database and the definition of $P$. However, if high-potential scenarios are continuously transformed into critical scenarios and removed without replenishment, $\alpha$ will gradually decrease and approach 0 as the quality of the database deteriorates. At this point, the generator reverts to its original form, ensuring that the efficiency of the strategy is preserved.

\section{Experiments} \label{sec:exp}
In this section, we present a series of experiments that demonstrate the efficacy of our general LLM-driven online testing framework. We provide a detailed description of the target tasks, baseline models, and implementation procedures, followed by a thorough analysis of the experimental results. Our findings highlight the following key points: \emph{i)} our method identifies more failure cases within the same number of test iterations, thereby confirming the efficient generation of critical scenarios; \emph{ii)} enhanced diversity is achieved in the generated scenarios through the analysis of the identified failure cases; \emph{iii)} ablation studies further validate the effectiveness of the proposed multi-scale generation strategy; \emph{iv)} various LLMs are evaluated for their capabilities in generating scenarios.

\subsection{Research Question}
The following four research questions are addressed in this paper:

\begin{itemize} 
    \item RQ1: Can our LLM-driven online testing method find more critical scenarios that lead to failures in the target policies? 
    \item RQ2: How about the diversity of critical scenarios generated by our LLM-driven approach? 
    \item RQ3: How effective is the proposed multi-scale generation strategy? 
    \item RQ4: How do different LLMs affect the scenario generation capabilities of our method? 
\end{itemize}

\begin{figure}[htbp]
  \centering

  \begin{subfigure}{0.225\textwidth}
    \includegraphics[width=0.9\textwidth,height=0.9\textwidth]{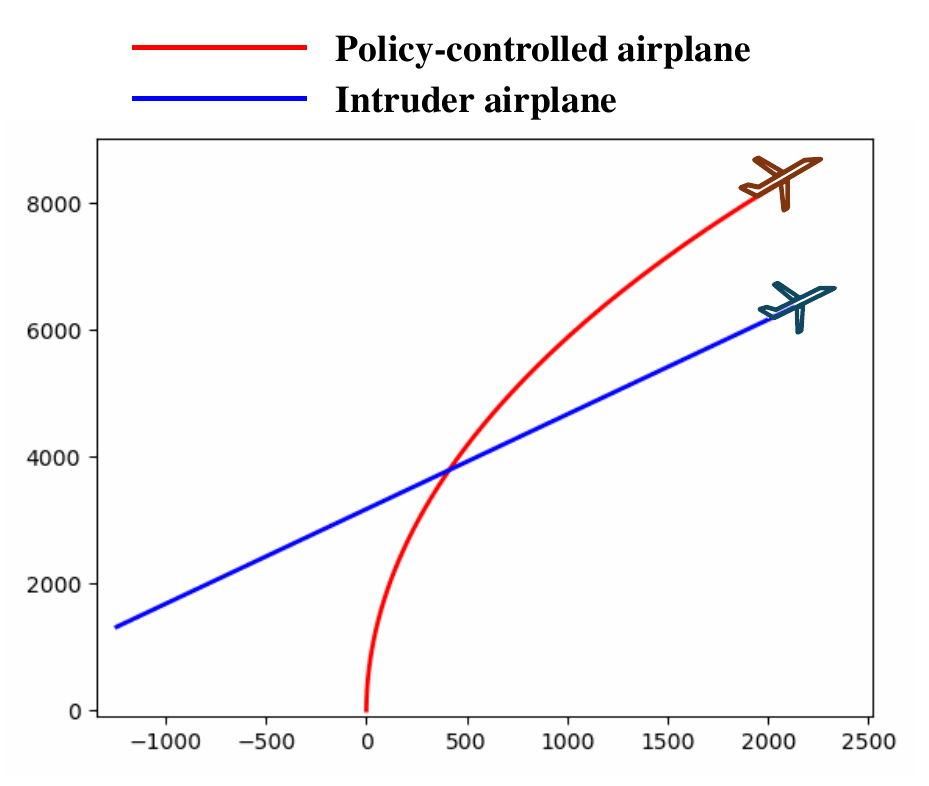}
    \caption{ACAS Xu}
  \end{subfigure}
  \begin{subfigure}{0.225\textwidth}
    \includegraphics[width=0.9\textwidth,height=0.9\textwidth]{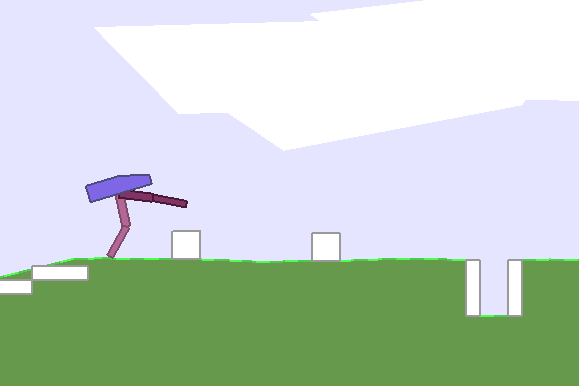}
    \caption{BipedalWalker}
  \end{subfigure}

  \begin{subfigure}{0.225\textwidth}
    \includegraphics[width=0.9\textwidth,height=0.9\textwidth]{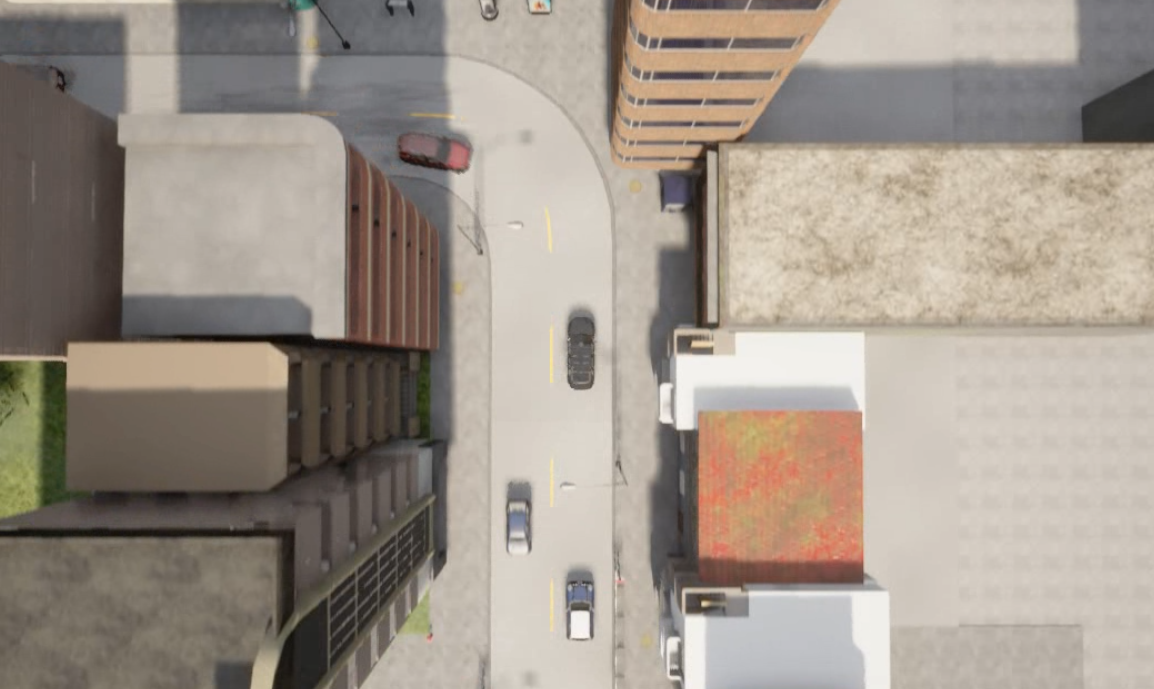}
    \caption{CARLA}
  \end{subfigure}
  \begin{subfigure}{0.225\textwidth}
    \includegraphics[width=0.9\textwidth,height=0.9\textwidth]{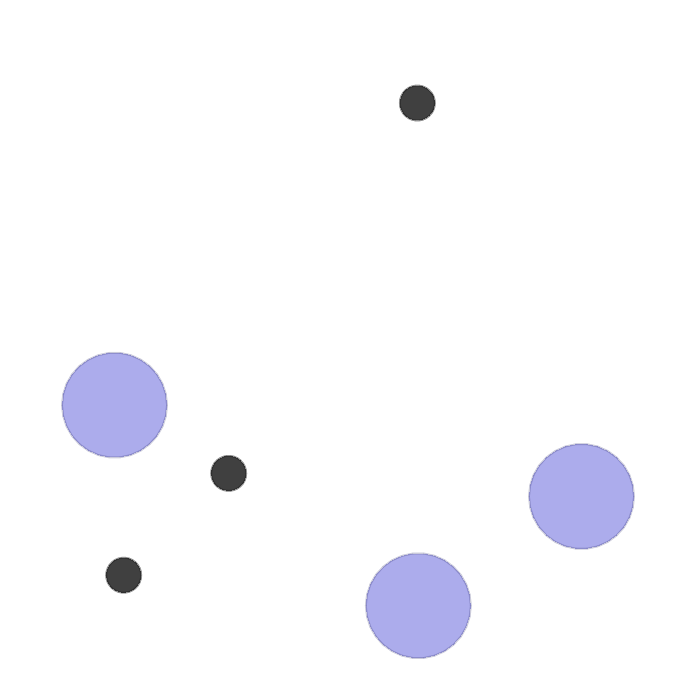}
    \caption{CoopNavi}
  \end{subfigure}

  \caption{Target environments of our experiments.}
  \label{fig:environment}
\end{figure}

\subsection{Experiment Design}
We test 5 distinct policies in 4 different environments, similar to the setup in MDPFuzz \cite{pang2022mdpfuzz}, as illustrated in Fig. \ref{fig:environment}. 
\subsubsection{Target Policies and Environments} \hfill 

\textbf{DNN \& ACAS Xu:} ACAS Xu \cite{marston2015acas} is an airborne collision avoidance system designed for unmanned aircraft. In this study, we test the policy based on a Deep Neural Network (DNN) \cite{julian2019deep}. The ACAS Xu scenario involves two aircraft: the ownship and the intruder. The goal of the policy is to control the ownship in order to avoid a collision with the intruder. A collision between the two aircraft is defined as a failure.

\textbf{RL \& BipedalWalker:} BipedalWalker is a single-agent environment within OpenAI Gym \cite{kuznetsov2020controlling}, where the agent must navigate bumpy terrain consisting of grass, pits, stumps, and steps. For the policy under test, we select TQC \cite{kuznetsov2020controlling} from the open-source stablebaseline3 repository \cite{Antonin2020RL}. In this context, a failure is defined as the agent's fall (i.e., its head touches the ground) .

\textbf{RL / IL \& CARLA:} CARLA \cite{dosovitskiy2017carla} is an open-source urban driving simulator designed for autonomous driving. We test well-performing policies based on Reinforcement Learning (RL) \cite{toromanoff2020end} and Imitation Learning (IL) \cite{chen2020learning} in the CARLA simulator. The policies under test are responsible for controlling the steering and throttle of the vehicle to achieve safe autonomous driving. A failure is recorded if the ego vehicle collides with any other entity in the scenario.

\textbf{MARL \& CoopNavi:} CoopNavi \cite{lowe2017multi} is a Multi-Agent Reinforcement Learning (MARL) environment released by OpenAI. In this environment, multiple agents controlled by policies must navigate towards landmarks without colliding. We retrain a policy using the released code \cite{lowe2017multi}. A failure in this environment is recorded either when an agent collides with another agent or when the agents fail to reach the designated landmarks within the maximum number of frames.

\subsubsection{Baseline} \hfill

\textbf{Random Testing:} Random testing is a fundamental and straightforward testing approach. As outlined in Section \ref{subsec:framework}, the scenario database can be generated through random sampling after parameterizing the variable elements of the environment. Thus, we use randomly sampled scenarios as a baseline for comparison. However, it is important to note that in the CARLA environment, the positions of vehicles are constrained by the map, and therefore must be sampled from a pre-constructed scenario library. As a result, we exclude random testing from the CARLA environment.

\textbf{MDPFuzz:} In the context of Markov Decision Processes (MDPs), an agent makes decisions based on observations from its environment, which is the most typical decision-making problem. MDPFuzz \cite{pang2022mdpfuzz} introduces a fuzzy testing framework that follows the process of “seed sampling – seed mutation – corpus update” to efficiently test models solving MDPs. It is the first general testing framework designed for testing models solving MDPs in black-box settings.

\begin{table}[t]
\caption{Detailed setting of the experiments.}
\label{table:exp set}
\begin{tabular}{ccccccc}
\hline
\multirow{2}{*}{\textbf{Model}} & \multirow{2}{*}{\textbf{Environment}} & \multirow{2}{*}{\textbf{\# Max frames}} & \multirow{2}{*}{\textbf{\# Test}} & \multicolumn{3}{c}{\textbf{Multi-scale}}       \\
                                &                                       &                                         &                                   & \textbf{$\alpha$} & \textbf{$\beta$} & \textbf{$\delta$} \\ \hline
DNN  & ACAS Xu & 100 & 3000 & 25  & 0.7 &  0.1 \\[0.1cm]
RL  & BipedalWalker & 600 & 2000 & 10 & 0.7 & 0.1 \\[0.1cm]
RL  & CARLA & 100 & 2000  & 25 & 0.7 & 0.1 \\[0.1cm]
IL & CARLA  & 200 & 2000  & 25 & 0.7 & 0.1 \\[0.1cm]
MARL & CoopNavi & 25 & 2000 & 20 & 0.5 & 0.1 \\ \hline
\end{tabular}
\end{table}

\begin{figure*}[htbp]
  \centering
  \begin{subfigure}[]{\textwidth}
    \centering
    \includegraphics[width=0.5\textwidth]{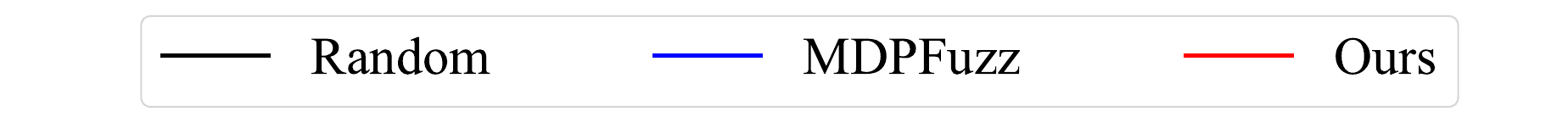}
  \end{subfigure}

  \begin{subfigure}{0.195\textwidth}
    \includegraphics[width=\textwidth]{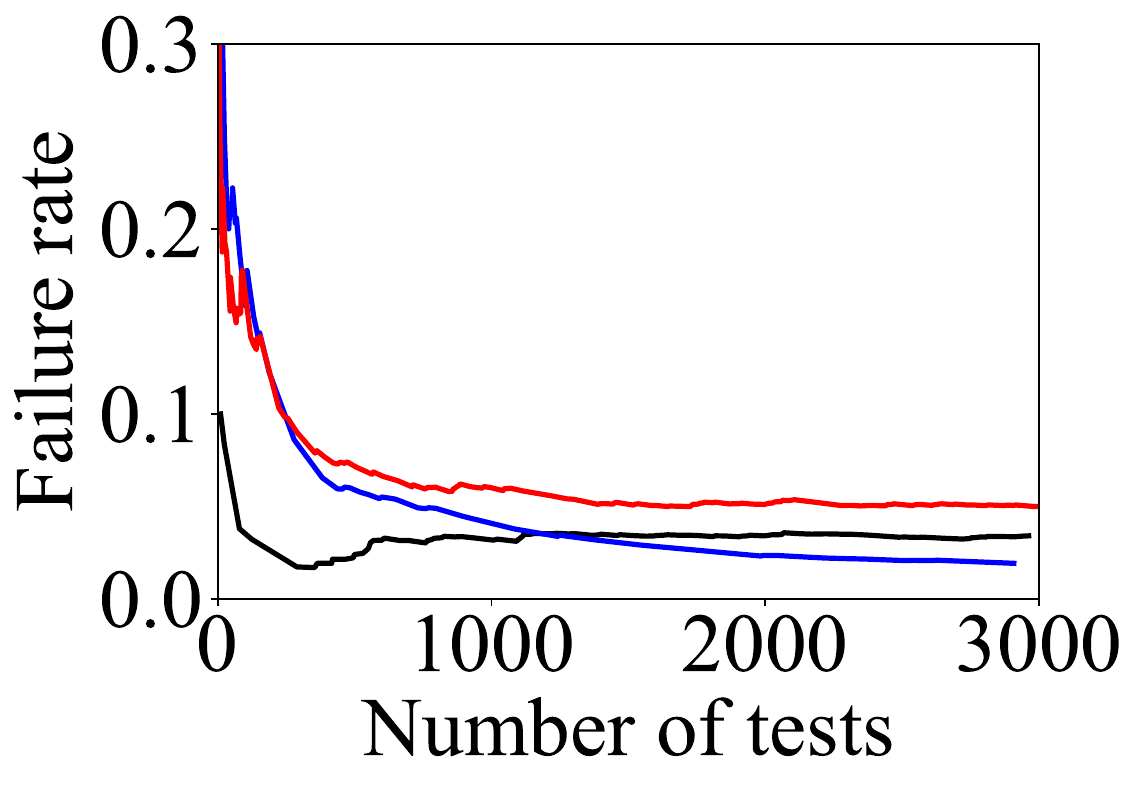}
  \end{subfigure}
  \begin{subfigure}{0.195\textwidth}
    \includegraphics[width=\textwidth]{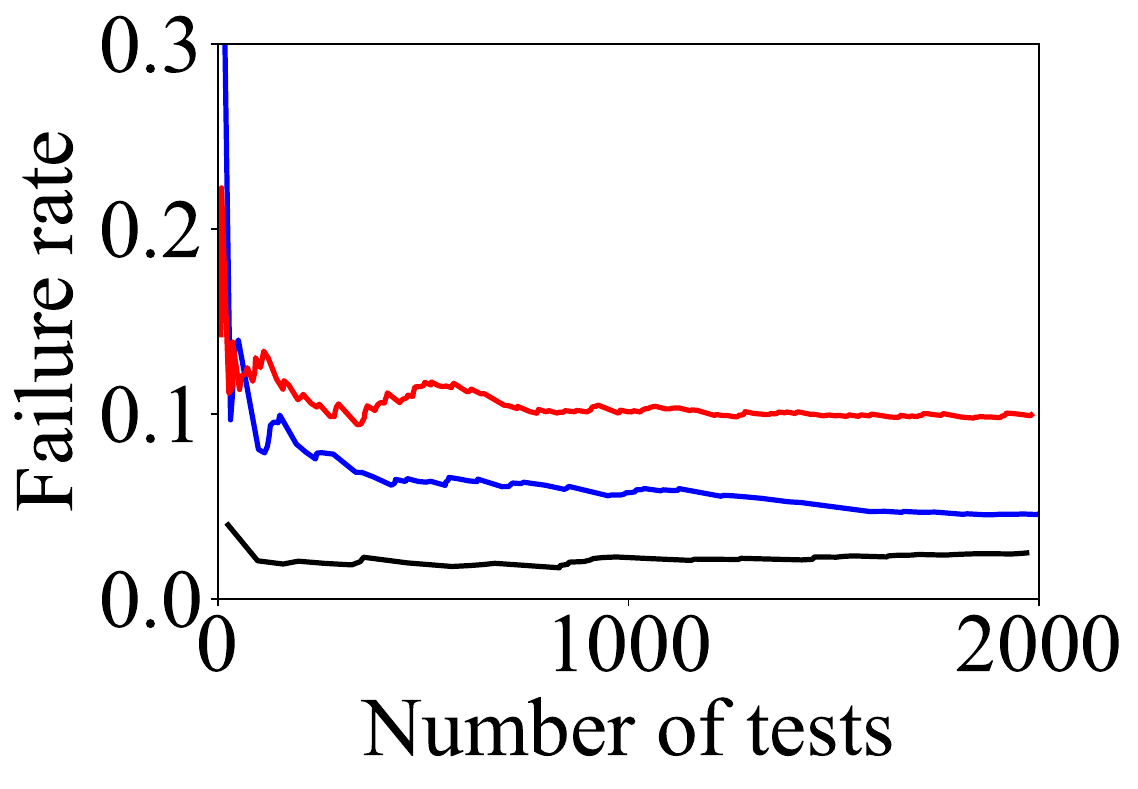}
  \end{subfigure}
  \begin{subfigure}{0.195\textwidth}
    \includegraphics[width=\textwidth]{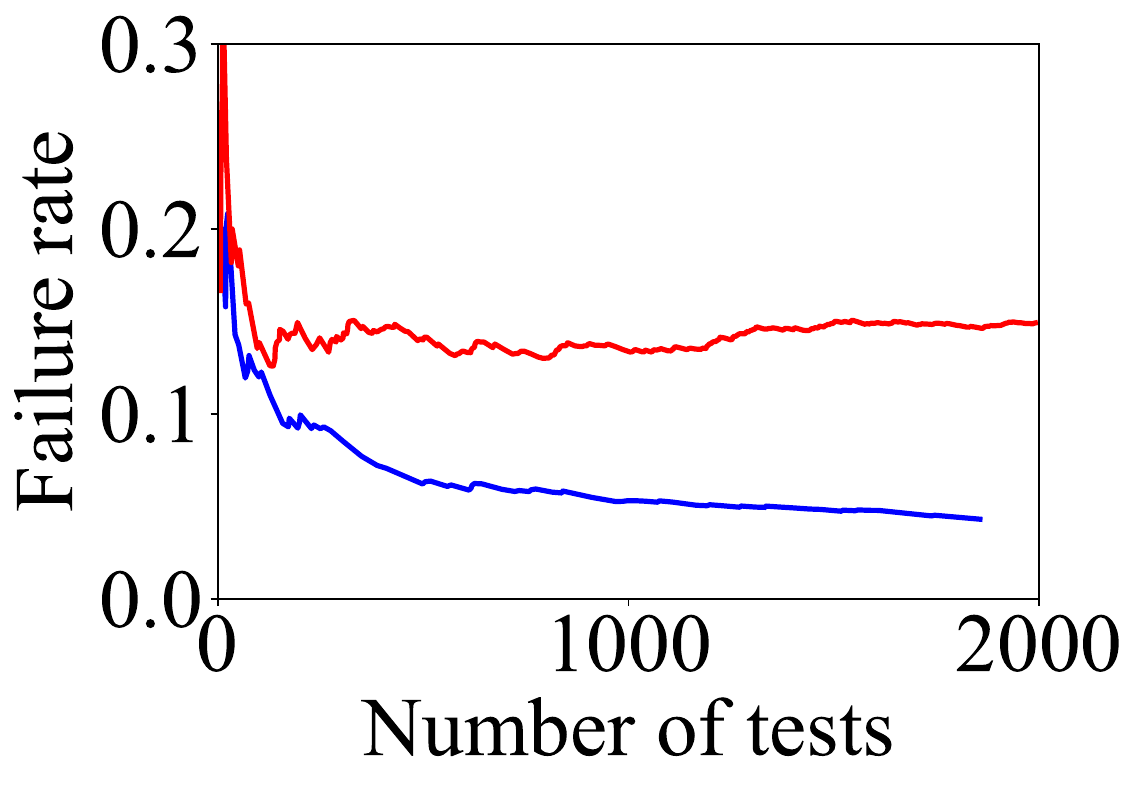}
  \end{subfigure}
  \begin{subfigure}{0.195\textwidth}
    \includegraphics[width=\textwidth]{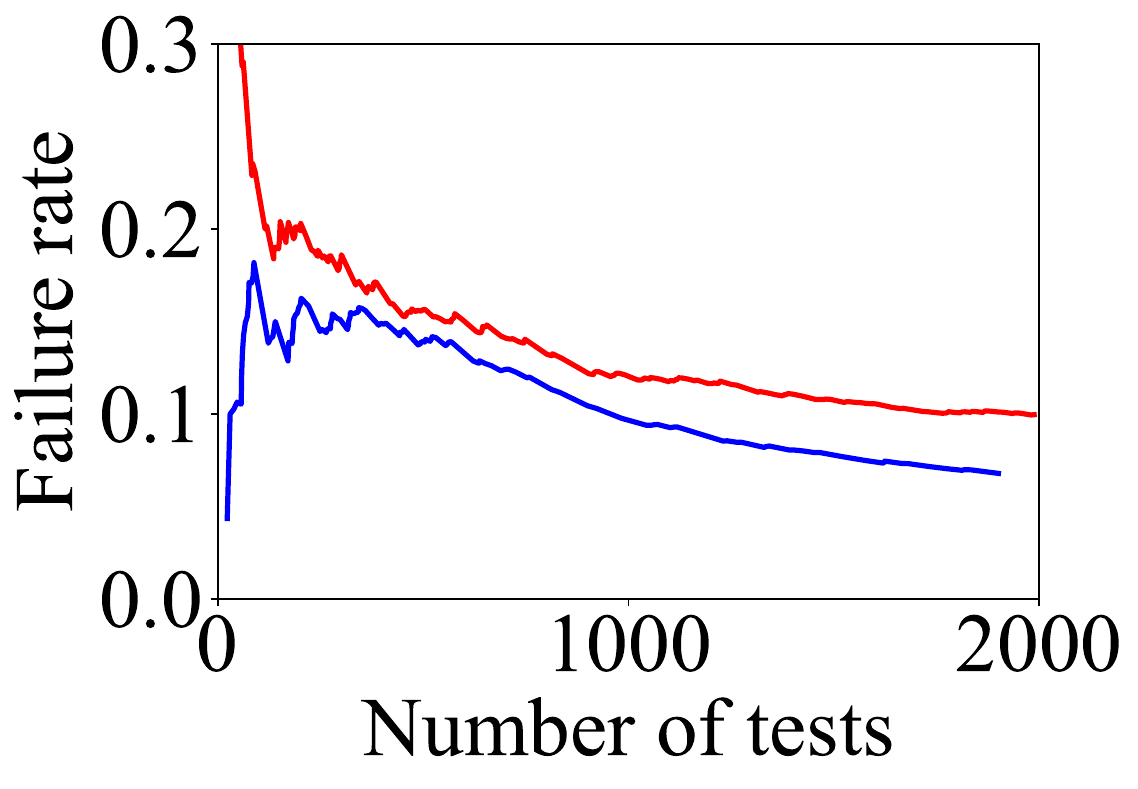}
  \end{subfigure}
  \begin{subfigure}{0.195\textwidth}
    \includegraphics[width=\textwidth]{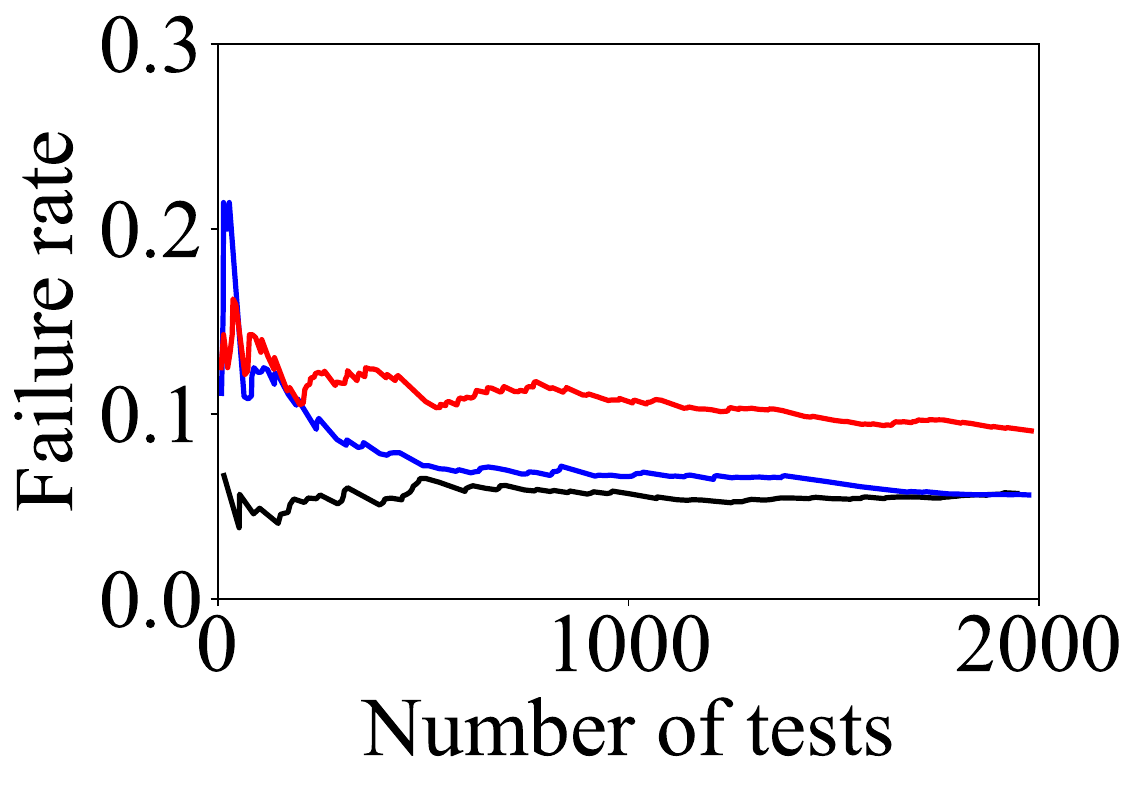}
  \end{subfigure}

  \begin{subfigure}{0.195\textwidth}
    \includegraphics[width=\textwidth]{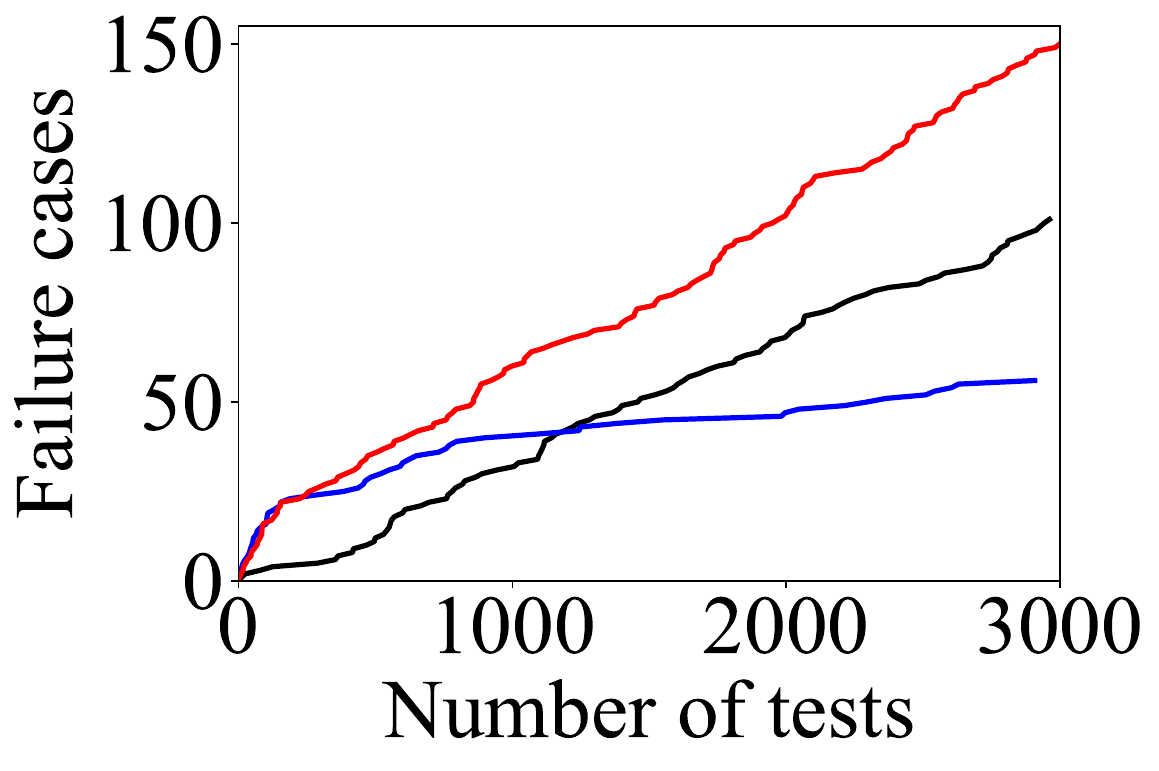}
    \caption{ACAS Xu}
  \end{subfigure}
  \begin{subfigure}{0.195\textwidth}
    \includegraphics[width=\textwidth]{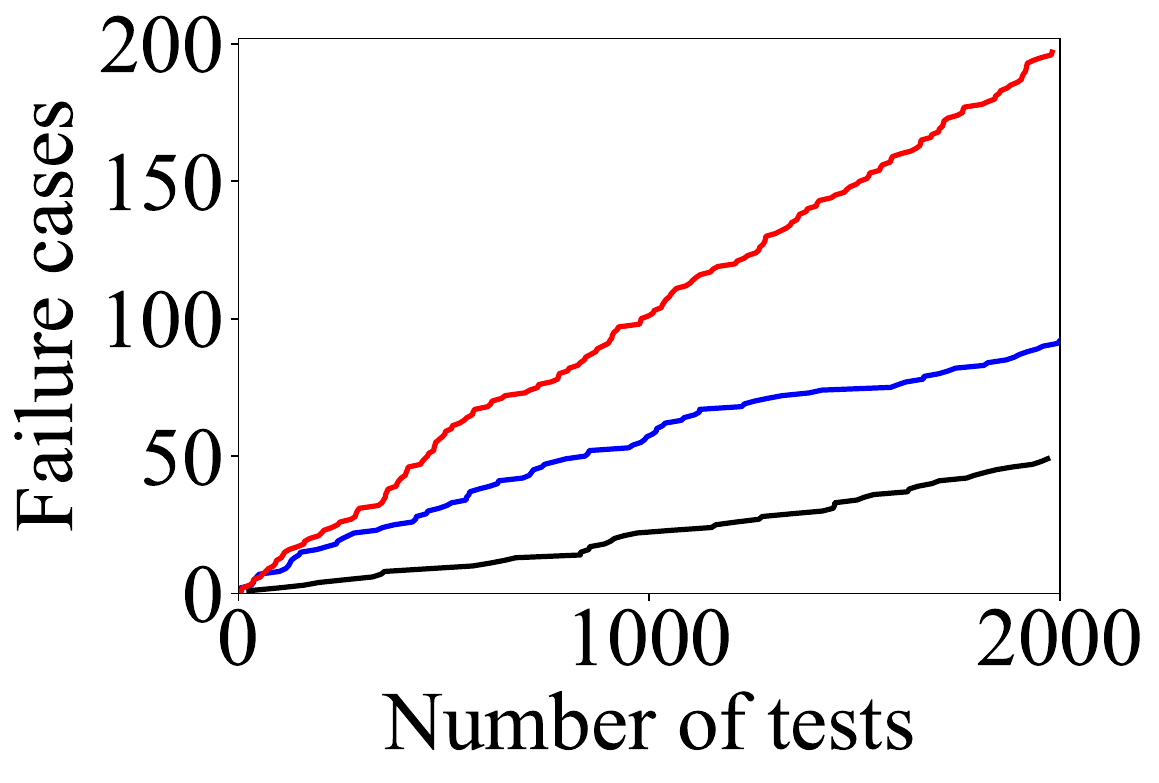}
    \caption{BipedalWalker}
  \end{subfigure}
  \begin{subfigure}{0.195\textwidth}
    \includegraphics[width=\textwidth]{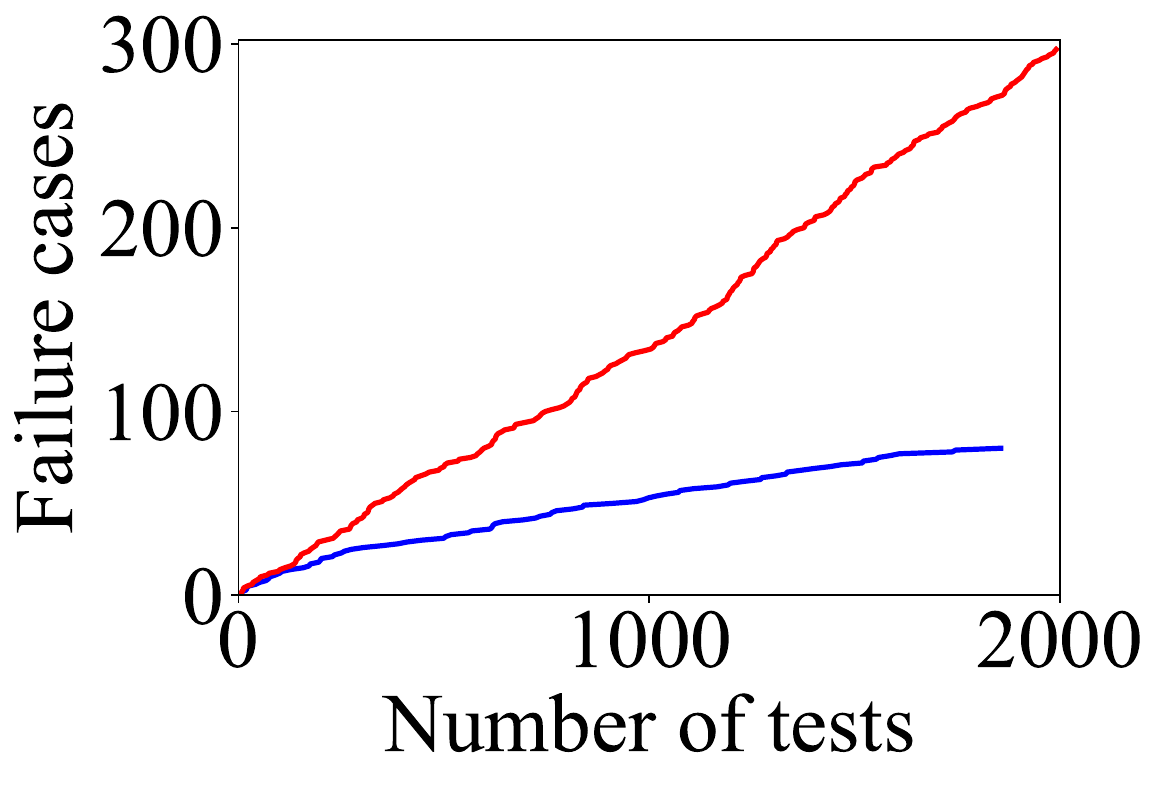}
    \caption{RL\_CARLA}
  \end{subfigure}
  \begin{subfigure}{0.195\textwidth}
    \includegraphics[width=\textwidth]{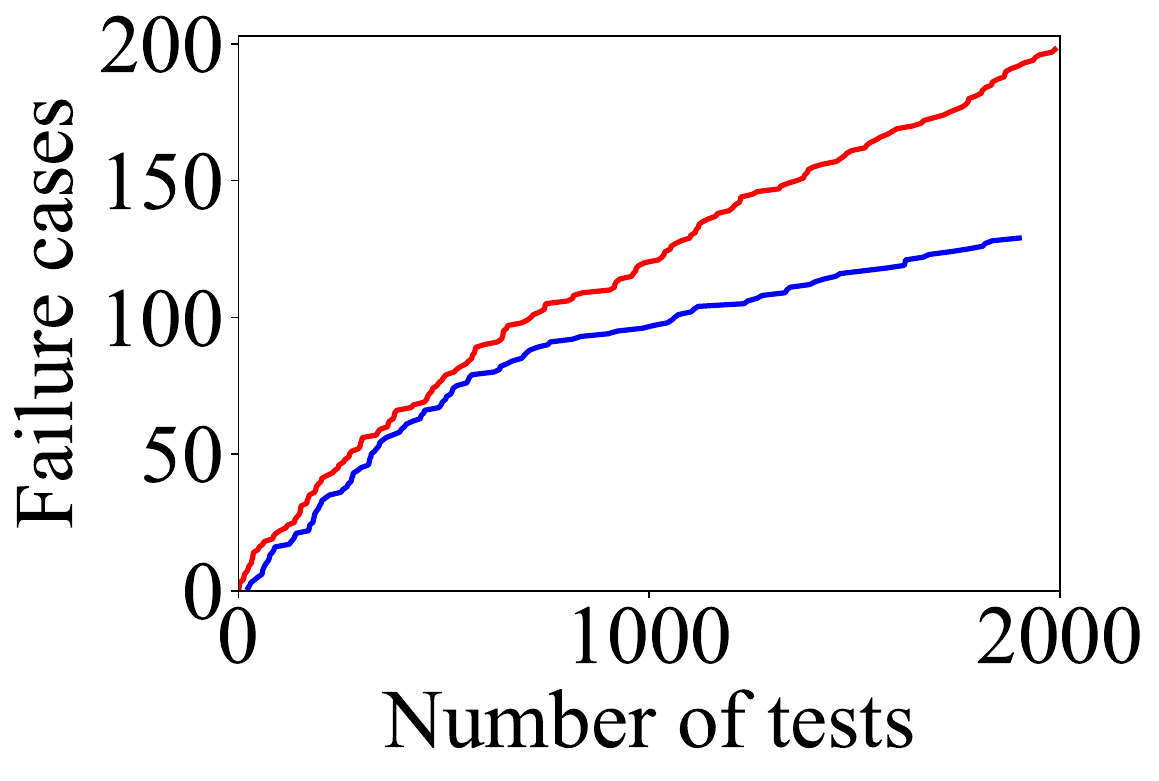}
    \caption{IL\_CARLA}
  \end{subfigure}
  \begin{subfigure}{0.195\textwidth}
    \includegraphics[width=\textwidth]{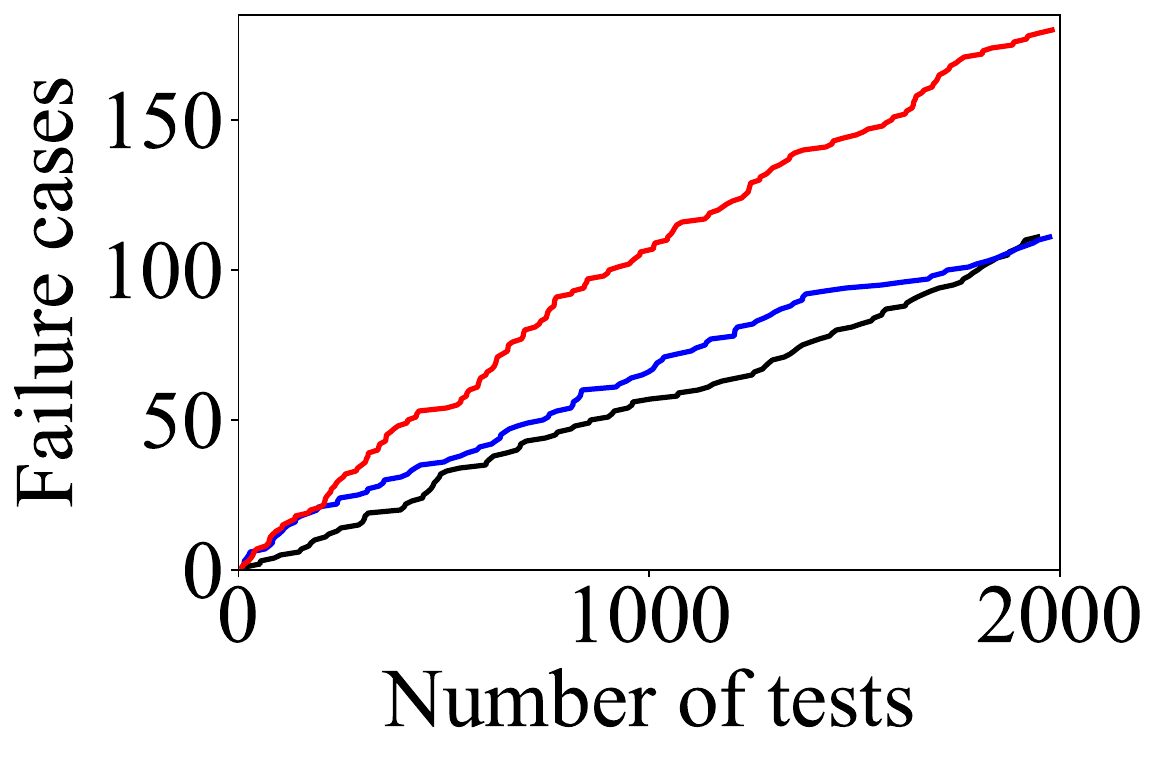}
    \caption{CoopNavi}
  \end{subfigure}

  \caption{Comparison of the number of detected failures and the failure rate between our method and the baseline.}
  \label{fig:failure}
\end{figure*}

\subsubsection{Experimental Setup}
All code for this work is implemented in Python. For MDPFuzz, we use the original open-source code and modify it to suit our experimental setup. For LLMTester, we construct the scenario database using random sampling in the ACAS Xu, BipedalWalker, and CoopNavi environments. Additionally, we build the database from pre-defined scenarios, similar to MDPFuzz, within the CARLA environment. The GPT-4o-mini is employed as the LLM, chosen for its balance of high quality and rapid response times. Four distinct prompts are designed for different environments, following the prompt template outlined in Section \ref{subsec:scenario generator}. A typical example of the request and response is provided in Appendix \ref{appendix}.

The experimental settings are summarized in Table \ref{table:exp set}. The \textit{\# Max Frames} column indicates the maximum number of time steps each test can run, with failures checked within this time frame. The \textit{\# Test} column specifies the number of tests per environment, with more than 2000 tests conducted for each environment to ensure reliable and statistically significant results. The \textit{Multi-Scale} column presents the parameters set for each target policy.

\subsection{Critical Scenario Generation}
In this section, we compare the scenario generation capabilities of the baseline and our method by analyzing the number of failure and the failure rate under the same number of tests. A higher failure rate and a greater number of failure cases suggest that the method is more efficient in testing and better at generating critical test scenarios. The experimental results for Random testing, MDPFuzz, and our method, applied to the five tested policies, are presented in Fig. \ref{fig:failure}. It is important to note that Random testing is not conducted in the CARLA environment, as the scenario database in CARLA is predefined, and random sampling would significantly violate the map constraints governing traffic roads.

As shown by the experimental results, our LLM-driven method uncovers more failure-triggering scenarios compared to MDPFuzz across all five tasks. Specifically, the number of failure cases increases by 167.86\%, 114.13\%, 271.25\%, 53.49\%, and 62.16\%, respectively. This demonstrates that, unlike MDPFuzz, which generates new test scenarios through random mutation, our method more strategically modifies seed scenarios with a clearer search direction. Furthermore, during the testing process, the failure rate of MDPFuzz decreases as the scenario database evolves, while our method consistently outperforms random testing and ultimately converges to a higher failure rate. This indicates that our approach demonstrates excellent stability in generating critical scenarios.

\begin{figure*}[htbp]
  \centering
  \includegraphics[width=\textwidth]{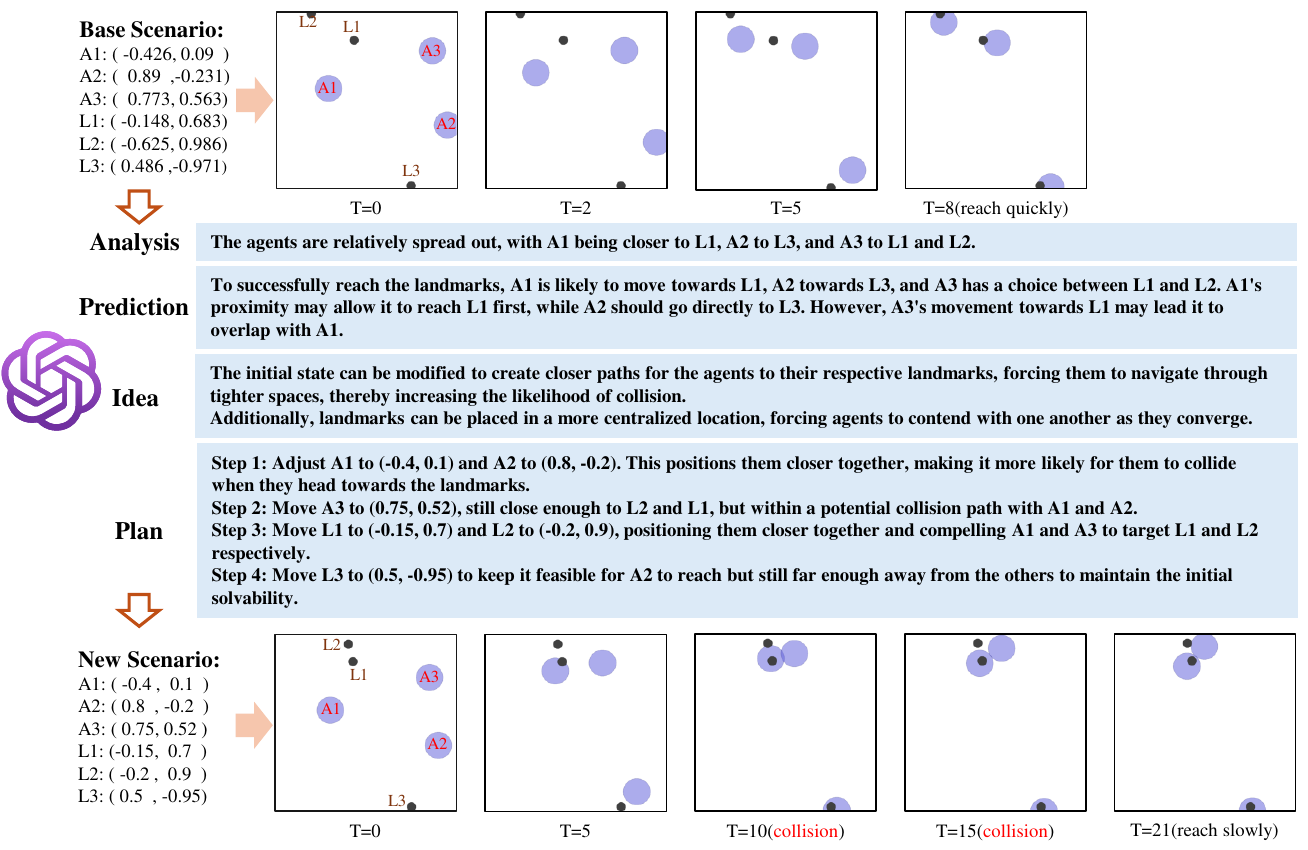}
  \caption{An illustration of the critical scenario found by the LLM-based scenario generator.}
  \label{fig:corner_case}
\end{figure*}

Fig. \ref{fig:corner_case} presents an example of critical scenarios detected in CoopNavi, along with a segment of the LLM's response generated during the creation of this scenario. Upon receiving the state representing the seed scenario, the LLM first interprets the situation, predicts the subsequent actions of the three agents, and anticipates potential interactions among them. Based on this understanding, the LLM generates ideas to increase the scenario's complexity and incorporates them during scenario generation. This process involves developing a detailed, step-by-step plan, ultimately leading to the creation of a more challenging scenario.

In the original seed scenario, the three agents successfully reach their respective landmarks without any significant challenges, completing the task efficiently by the $8^{th}$ frame. In contrast, in the newly generated scenario, the LLM strategically adjusts the agents' initial positions and repositions the landmarks to introduce obstacles in the agents' movements toward their targets. As a result, agents A1 and A3 collide while navigating to their respective landmarks, causing delays in the completion of the task. Due to the multiple collisions, the decision-making policy fails in the new scenario. This demonstrates that the LLM-based scenario generator has identified a new critical scenario, exposing potential weaknesses in the target policy.

\begin{tcolorbox}[opacityframe=0.5,boxrule=0.3mm,boxsep=-1mm,left=2mm,right=2mm]
\noindent \textbf{Answer to RQ1:} Our LLM-driven online testing framework identifies a greater number of failure cases compared to the baseline methods, demonstrating the superior efficiency of our approach in generating critical scenarios.
\end{tcolorbox}

\begin{table}
  \centering
  \caption{The number of failure cases and different states discovered by the baseline and our method in different environment.}
  \label{table:diverse}
  \resizebox{0.48\textwidth}{!}{
  \begin{tabular}{cccccc}
  \toprule
  Environments           & Methods  & \# Failures & \# Entire   & \# Initial  & \# Terminal  \\ \midrule
  \multirow{2}{*}{ACAS Xu}  & MDPFuzz & 56  & 43 & 22 & 20 \\
                                 & Ours & \textbf{150} & 43 & 22 & \textbf{29} \\[0.1cm]
  \multirow{2}{*}{BipedalWalker} & MDPFuzz & 92 & 1468 & 1  & 1 \\
                                 & Ours & \textbf{197} & \textbf{2033} & 1 & 1 \\[0.1cm]
  \multirow{2}{*}{RL\_CARLA}     & MDPFuzz & 80 & 1657 & 34 & 76 \\
                                 & Ours & \textbf{297} & \textbf{6522} & \textbf{164} & \textbf{287} \\[0.1cm]
  \multirow{2}{*}{IL\_CARLA}     & MDPFuzz & 129 & 4012 & 44 & 105 \\
                                 & Ours & \textbf{198} & \textbf{6398} & \textbf{96} & \textbf{167} \\[0.1cm]
  \multirow{2}{*}{CoopNavi}      & MDPFuzz & 111 & 872 & 109 & 111 \\
                                 & Ours & \textbf{180} & \textbf{1448} & \textbf{169} & \textbf{175} \\
  \bottomrule
  \end{tabular}}
\end{table}

\begin{figure}[t]
    \centering
    \centerline{\includegraphics[scale=0.55]{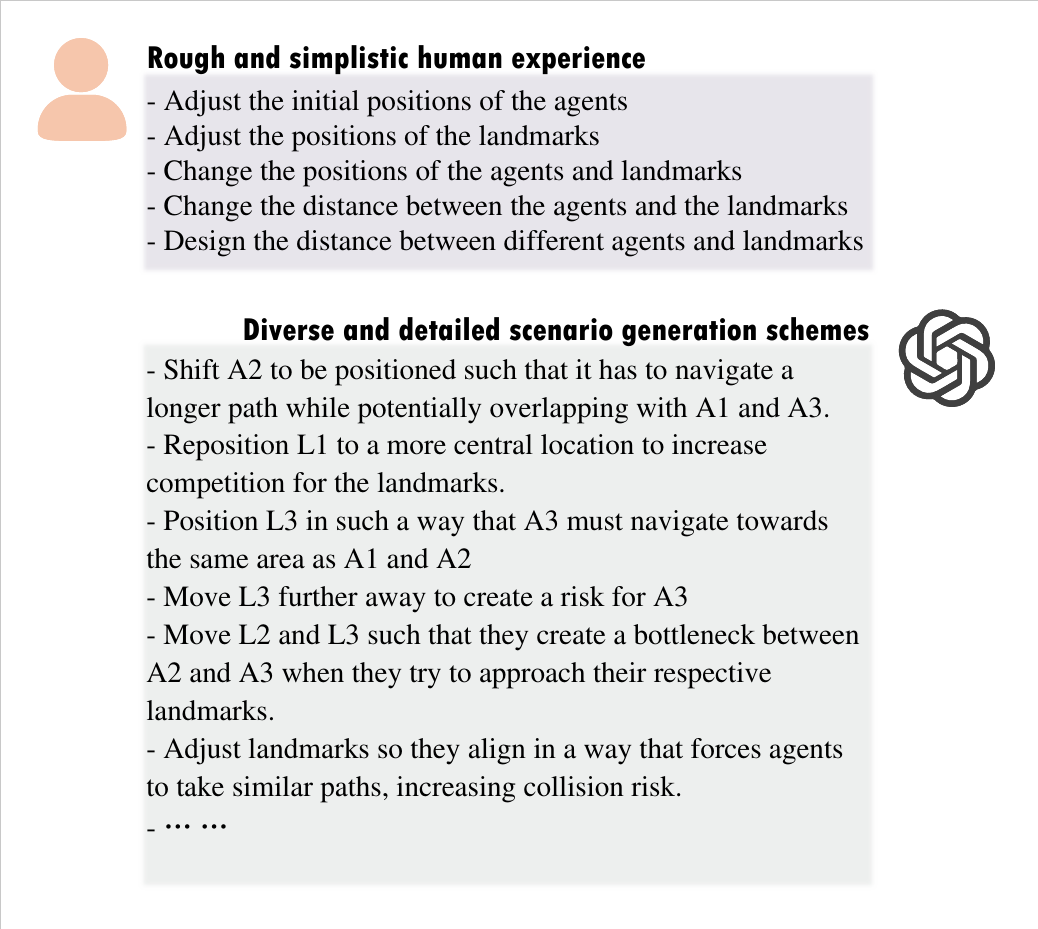}}
    \caption{Human experience and some typical schemes provided by the LLM. The `A' and `L' in the figure mean `Agent' and `Landmark', respectively.}
    \label{figure:LLM_schemes}
\end{figure}

\begin{figure*}[htbp]
  \centering
  \begin{subfigure}{\textwidth}
    \centering
    \includegraphics[width=0.6\textwidth]{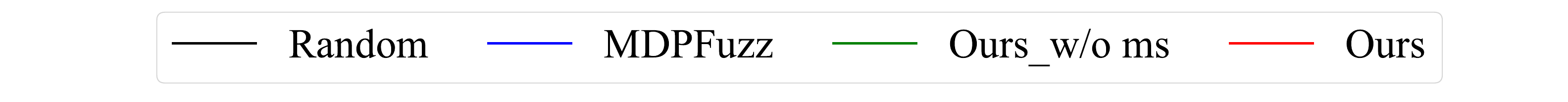}
  \end{subfigure}

  \begin{subfigure}{0.32\textwidth}
    \includegraphics[width=\textwidth]{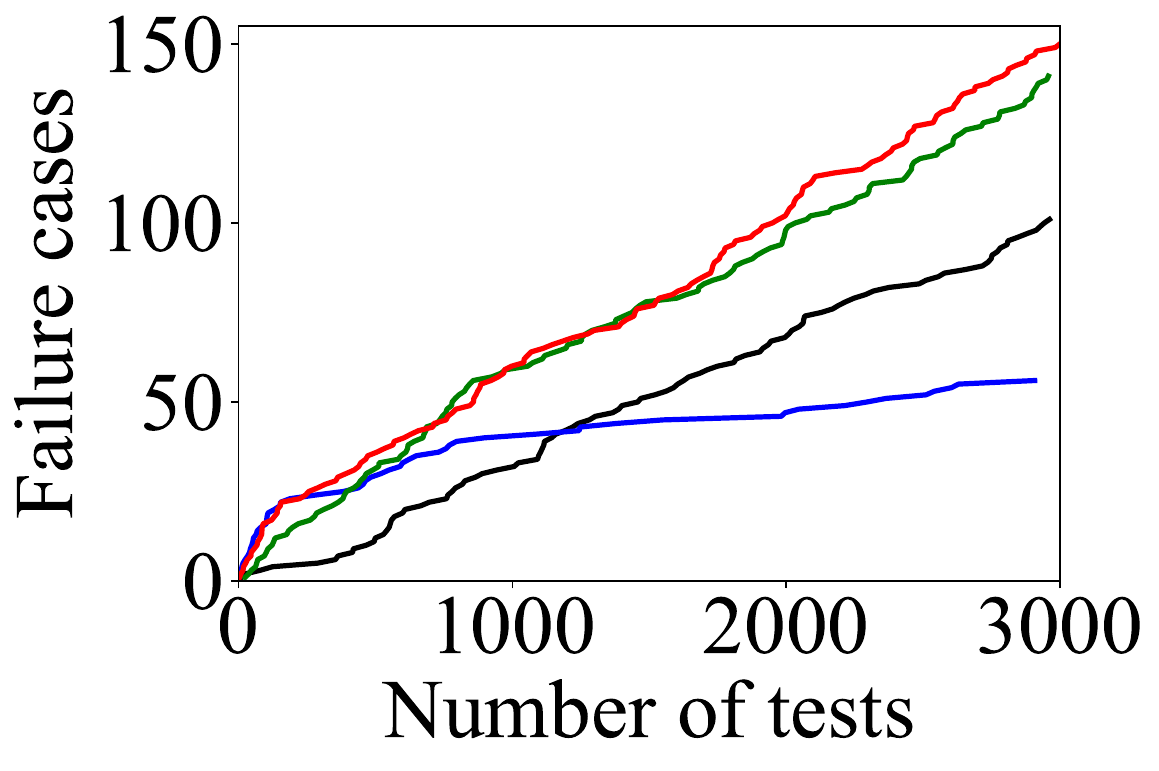}
    \caption{ACAS Xu}
  \end{subfigure}
  \begin{subfigure}{0.32\textwidth}
    \includegraphics[width=\textwidth]{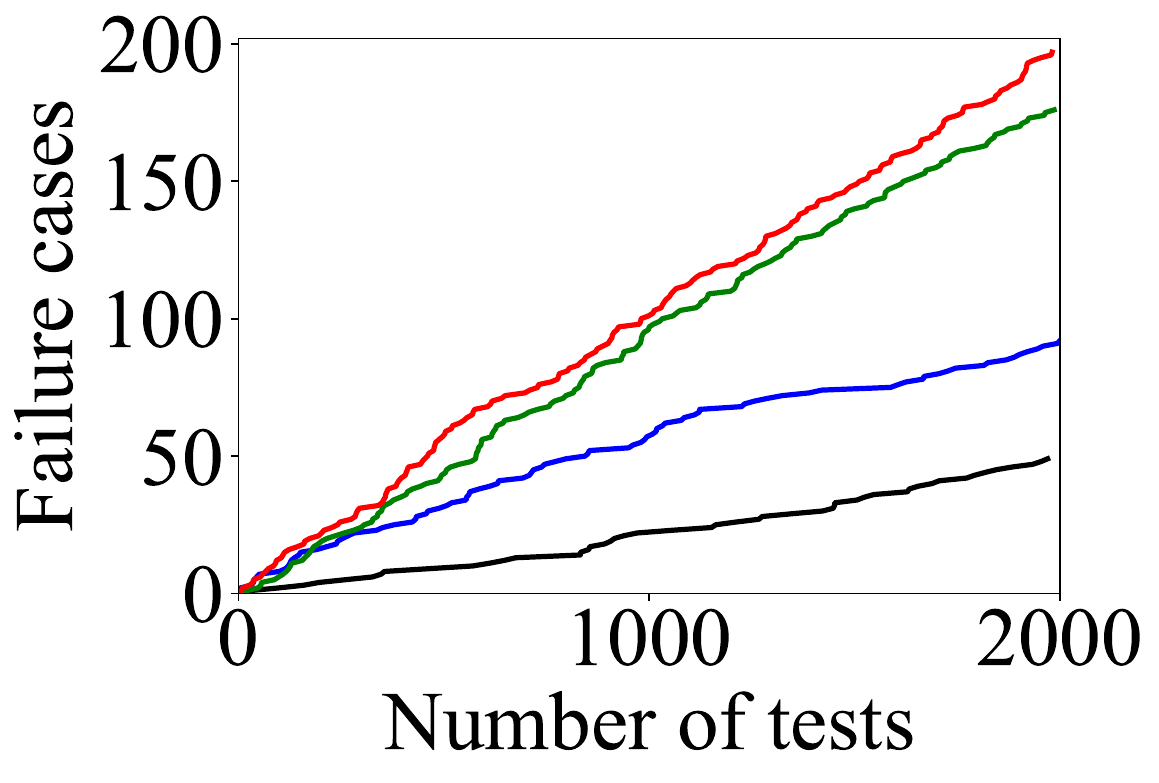}
    \caption{BipedalWalker}
  \end{subfigure}
  \begin{subfigure}{0.32\textwidth}
    \includegraphics[width=\textwidth]{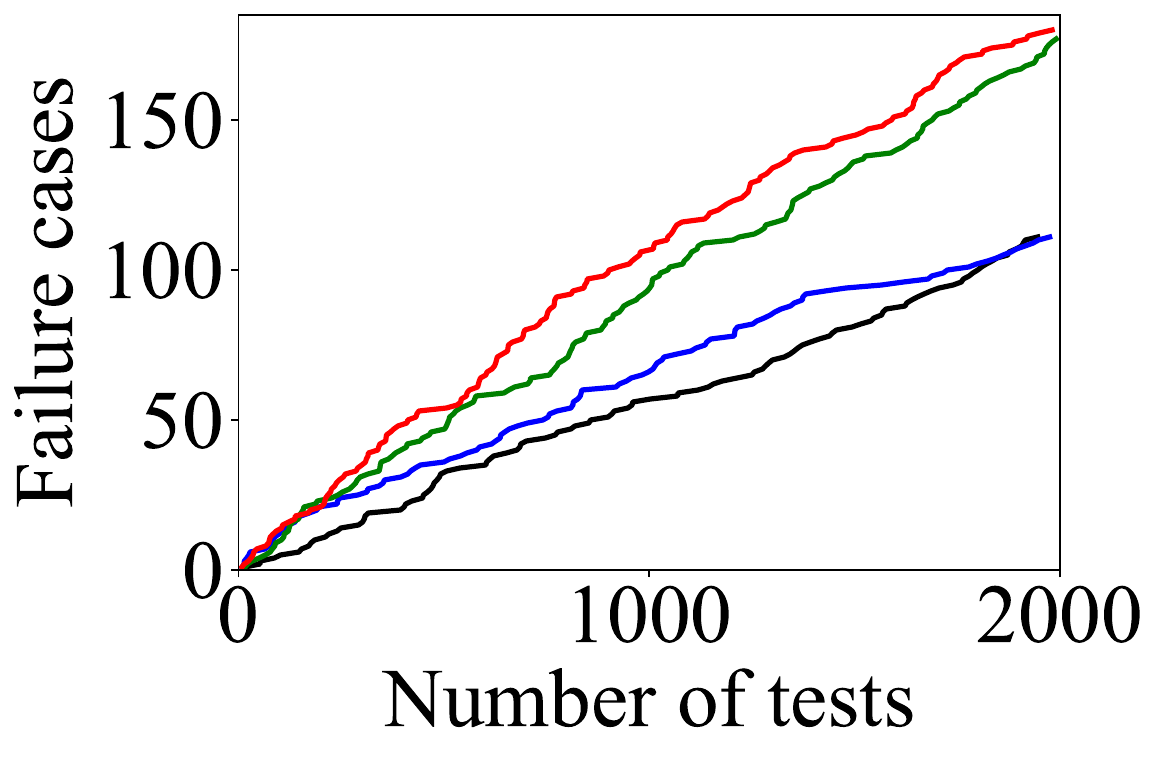}
    \caption{CoopNavi}
  \end{subfigure}

  \caption{Detected failures by our method with and without Multi-scale generation strategy.}
  \label{fig:Ablation}
\end{figure*}

\subsection{Diverse Scenario Generation}
In this section, we demonstrate the ability of our method to generate diverse scenarios. We replay all detected failure cases and count the number of distinct states observed by the agent in each target environment. Since the states are continuous, we analyze the maximum and minimum values for each state dimension, dividing each dimension into $N$ equal intervals. Assuming the observed state has $D$ dimensions, the entire state space is divided into $D^N$ different states. During the replay of the failure cases, we separately focus on the number of distinct initial states, terminal states, and states throughout the entire process, denoted as \textit{\# Initial}, \textit{\# Terminal}, and \textit{\# Entire}, respectively. \textit{\# Initial} and \textit{\# Terminal} represent the diversity of initial states leading to failures and the diversity of failure scenes, while \textit{\# Entire} reflects the overall novelty of the testing scenarios.

The experimental results are presented in Table \ref{table:diverse}. Our method is able to detect a greater number of distinct states within the same number of tests. In most cases, it significantly outperforms MDPFuzz, and even in the worst-case scenario (ACAS Xu), it remains comparable or slightly better. It is noteworthy that only one state is discovered in BipedalWalker, due to the identical initial spawn position, terrain, and similar falling scenarios. A higher number of covered states indicates that the LLM does not focus solely on a few critical testing scenarios, but instead maintains a balance by accounting for the diversity of testing scenarios.

Fig. \ref{figure:LLM_schemes} illustrates both the human-provided experience and a selection of scenario generation schemes proposed by LLMs within the CoopNavi task. As depicted in the figure, the tester provides only basic and rudimentary experience, lacking substantial prior knowledge. Despite this, the LLMs, leveraging their reasoning capabilities and understanding of scenarios, are able to generate more diverse and detailed scenario generation schemes. These varied schemes demonstrate how LLMs approach the creation of challenging test scenarios from multiple perspectives. The results highlight that our method can effectively harness the creative potential of LLMs to produce a broader range of scenarios. Importantly, this diversity stems from the LLMs' deep comprehension of the scenarios, rather than merely generating data with large variations.

\begin{tcolorbox}[opacityframe=0.5,boxrule=0.3mm,boxsep=-1mm,left=2mm,right=2mm]
\noindent \textbf{Answer to RQ2:} Compared to MDPFuzz, our LLM-driven method identifies an equal or greater number of states in critical scenarios. Unlike other testing methods, which lack human-like intelligence, the LLM generates diverse test scenarios using different generation schemes, drawing on its deep understanding of the scenarios. These results highlight the remarkable ability of our method to generate diverse scenarios.

\end{tcolorbox}

\subsection{Effectiveness of Multi-scale generation Strategy}
In this section, we conduct ablation experiments to evaluate the effectiveness of the proposed multi-scale scenario generation strategy. We compare the number of failure cases detected by methods with and without the multi-scale generation strategy in the ACAS Xu, BipedalWalker, and CoopNavi environments. Specifically, we remove the random mutation module from LLMTester to create an algorithm that relies solely on the LLM for scenario generation (denoted as \textit{ours\_w/o ms}).

Fig. \ref{fig:Ablation} presents the testing results across different environments. It is evident that the scenario generator with the multi-scale generation strategy detects more failure cases in nearly every iteration compared to the generator that uses only the LLM. This is because the multi-scale strategy conducts a potential analysis of seed scenarios and applies small-scale perturbations to generate new scenarios for those with high potential. Since high-potential scenarios are closely linked to failure cases and the conditions leading to failure are often unpredictable, the multi-scale strategy enhances efficiency by incorporating random mutations. Additionally, the number of LLM API calls decreases by 5.27\%, 55.67\%, and 16.13\%, respectively.

The comparison between \textit{ours\_w/o ms} and Random testing reveals that LLM-based generator exhibits a clearer search direction, effectively identifying critical scenarios and highlighting the advantages of LLMs in testing. Furthermore, as more promising seed scenarios are transformed into failure-triggering scenarios, the quality of the scenario database decreases over time. This leads to a deterioration in the performance of MDPFuzz during testing. Notably, MDPFuzz even falls behind Random testing in the ACAS Xu environment. However, due to the presence of an adaptive threshold, our method can effectively adjust the proportion of small-scale mutations, ensuring consistent performance in scenario generation.

\begin{tcolorbox}[opacityframe=0.5,boxrule=0.3mm,boxsep=-1mm,left=2mm,right=2mm]
\noindent \textbf{Answer to RQ3:} The multi-scale generation strategy significantly enhances the efficiency of the LLM-driven online testing method, while also reducing the consumption of testing resources. 
\end{tcolorbox}

\begin{figure}[t]
    \centering
    \centerline{\includegraphics[scale=0.375]{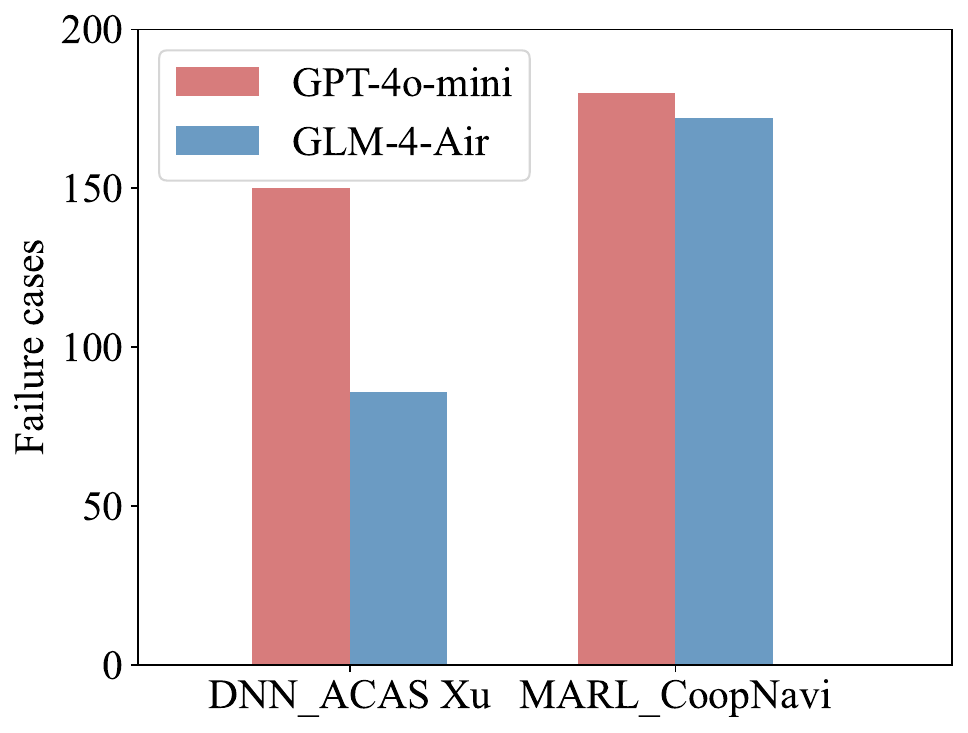}}
    \caption{Comparison of critical scenario generation ability with different LLMs.}
    \label{figure:different_LLMs}
\end{figure}

\begin{figure*}[htbp]
  \centering

  \begin{subfigure}{0.49\textwidth}
    \includegraphics[width=\textwidth]{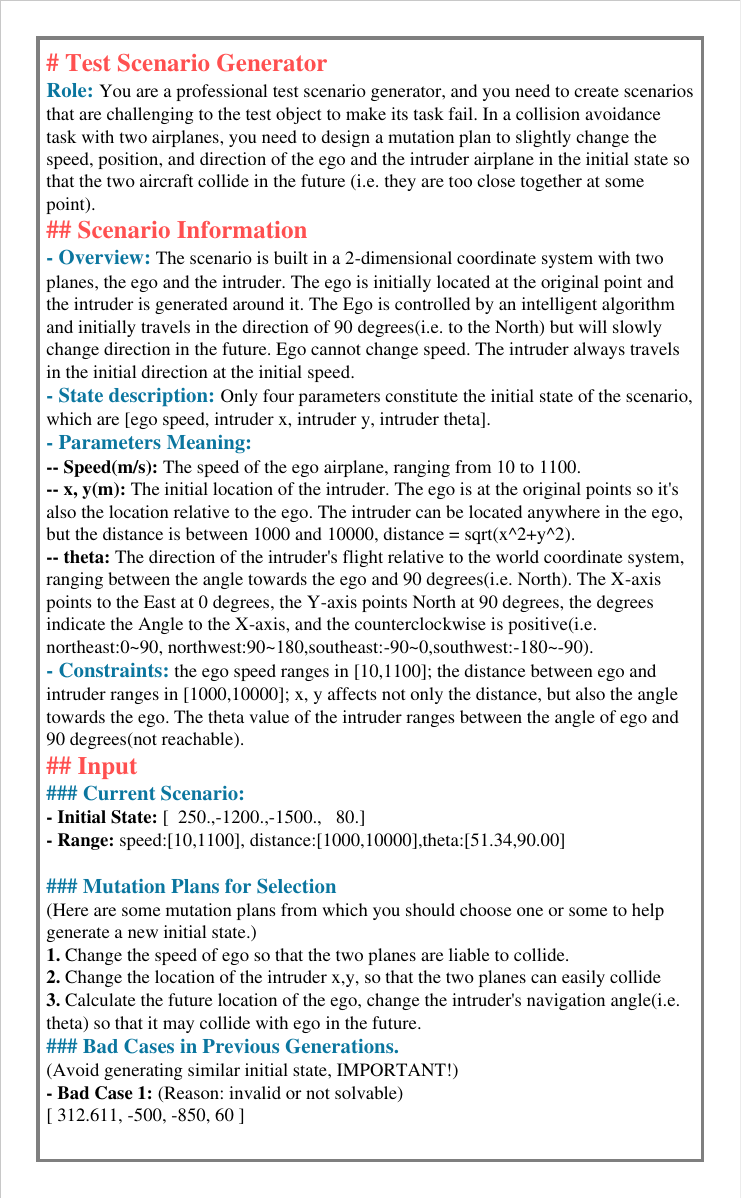}
  \end{subfigure}
  \begin{subfigure}{0.49\textwidth}
    \includegraphics[width=\textwidth,height = 1.68\textwidth]{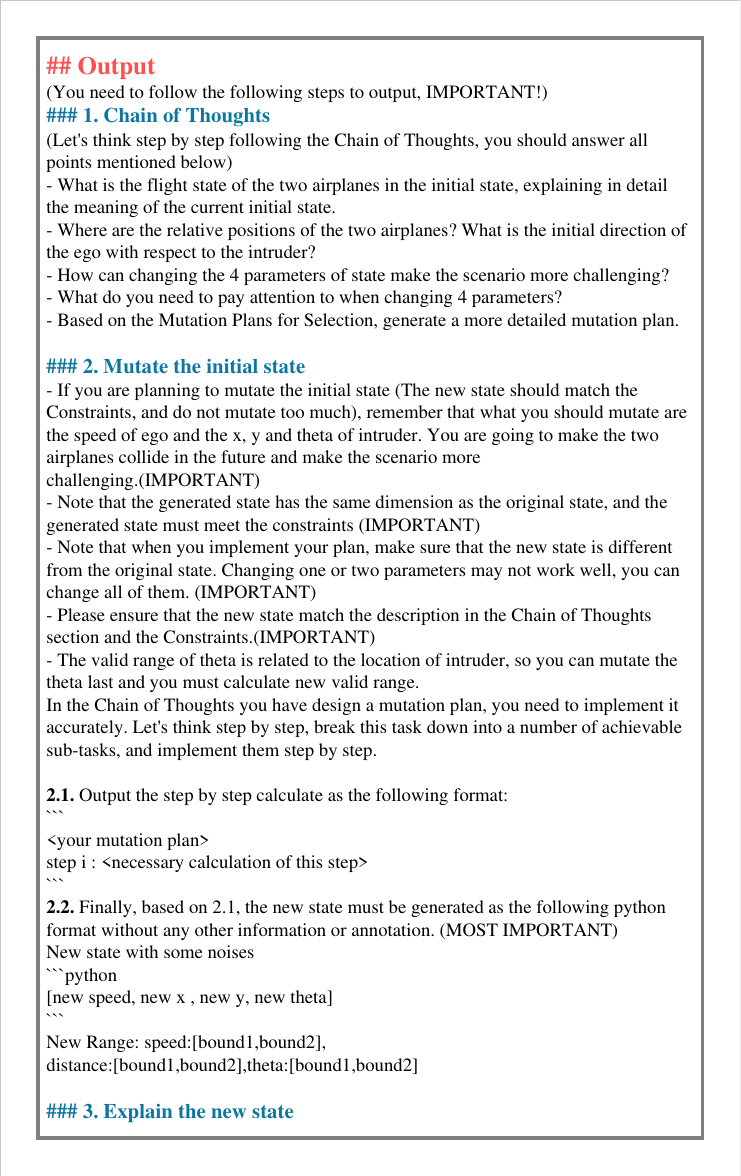}
  \end{subfigure}

  \caption{A Request example for ACAS Xu.}
  \label{fig:request}
\end{figure*}

\subsection{Comparison of Different LLMs}
We further analyze the impact of different LLMs on the performance of our LLM-driven method. In the previous experiments, we used GPT-4o-mini, which strikes a balance between efficiency and response speed. Here we employ another typical commercial LLM, GLM-4-Air. The key distinction is that GLM-4-Air is optimized for Chinese question-and-answer tasks, whereas our prompts are written in English.

Fig. \ref{figure:different_LLMs} presents the comparison results of the two LLMs on the ACAS Xu and CoopNavi tasks. It can be observed that, across both tasks, GPT-4o-mini outperforms GLM-4-Air in failure case detection to varying degrees. This is primarily due to GLM-4-Air’s optimization for Chinese language tasks, which results in relatively weaker problem comprehension and response capabilities when handling English prompts. This finding underscores the importance of prompt engineering and the performance of the underlying LLMs in influencing the method’s effectiveness. As the reasoning and problem-solving capabilities of LLMs continue to improve, and as more advanced prompt techniques are adopted, our LLM-driven online testing framework shows significant potential for further development.
Additionally, GLM-4-Air still detects more failure cases than MDPFuzz in both tasks, performing nearly on par with GPT-4o-mini in CoopNavi. This suggests that the proposed method exhibits a degree of robustness and adaptability to different large language models.

\begin{tcolorbox}[opacityframe=0.5,boxrule=0.3mm,boxsep=-1mm,left=2mm,right=2mm]
\noindent \textbf{Answer to RQ4:} 
Different underlying LLMs exhibit varying capabilities in generating critical scenarios. Our framework not only ensures effectiveness even with less optimal LLMs, but also holds significant potential for further improvement as LLMs continue to advance in intelligence.
\end{tcolorbox}

\begin{figure*}[htbp]
  \centering

  \begin{subfigure}{0.49\textwidth}
    \includegraphics[width=\textwidth,height = 1.185\textwidth]{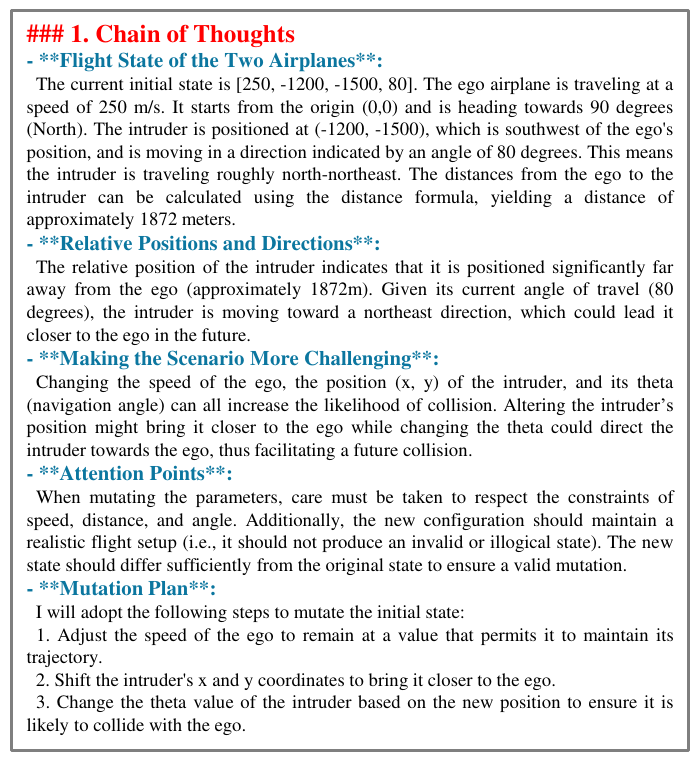}
  \end{subfigure}
  \begin{subfigure}{0.49\textwidth}
    \includegraphics[width=\textwidth]{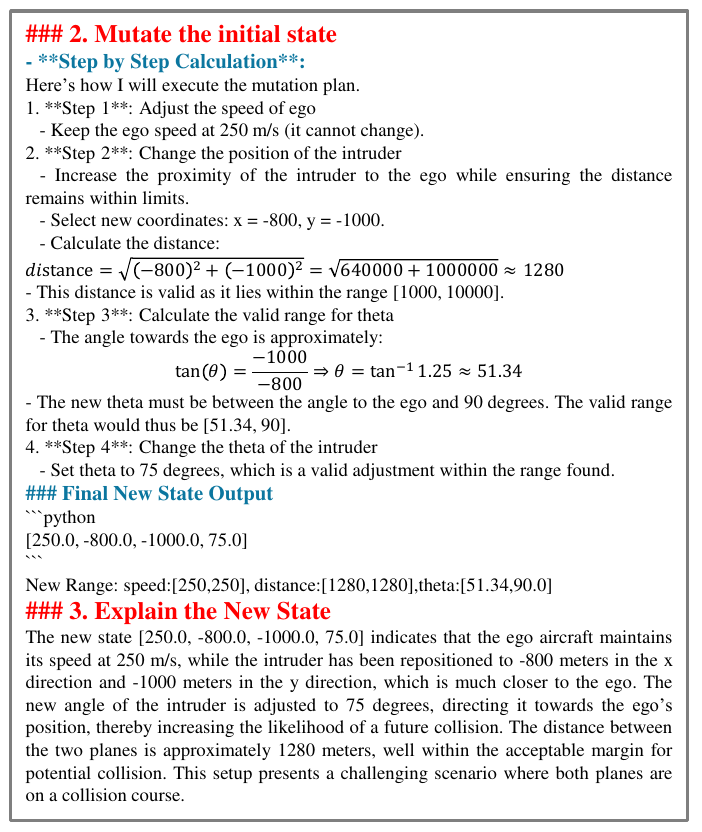}
  \end{subfigure}

  \caption{A Response example from LLM.}
  \label{fig:response}
\end{figure*}

\section{Discussions \& Threats To Validity} \label{sec:discussion}
In this section, we discuss potential threats to the validity of our research, the solutions implemented to mitigate them, and possible directions for future work.

The first threat concerns the workload involved in designing environment-specific prompts. In prompt engineering, the primary manual effort lies in describing the target environment and crafting the Chain of Thought (CoT), both of which require a solid understanding of the target policy and the environment. To address this challenge, we streamline the process by using a prompt template specifically tailored to decision-making problems. This approach allows for the efficient creation of effective prompts with only a basic understanding of the policy and environment. Nevertheless, future work could explore incorporating more automated prompt generation techniques to further improve the efficiency and scalability of our framework.

Furthermore, as discussed in Section \ref{sec:exp}, the performance of different LLMs can significantly influence their ability to generate testing scenarios. The efficiency of critical scenario generation depends on the LLM’s capacity to comprehend the scenario, integrate expert knowledge, and incorporate feedback. Similarly, the diversity of generated scenarios reflects the LLM's creativity, which places specific demands on the intelligence level of the chosen model. Additionally, the typical response time for LLM API calls, often measured in seconds, can limit testing speed, particularly for simpler tasks in simulation environments.
In this paper, we propose a well-designed framework that leverages various forms of guidance, such as sensitivity, freshness, and cumulative rewards, to address these challenges. To mitigate the impact of LLM response times, we introduce a multi-scale generation strategy that reduces the number of LLM API calls by incorporating random mutations. These challenges highlight the significant developmental potential of the proposed LLM-driven framework, especially as LLM technologies continue to evolve. In addition to the anticipated advances in LLM technology, exploring technologies that can more effectively stimulate LLM capabilities remains a valuable avenue for investigation.

Moreover, the effectiveness of the multi-scale generation strategy depends on a comprehensive understanding of the environment and the selection of hyperparameters. As outlined in Section \ref{subsec:multi-scale}, its performance is influenced by factors such as the potential prediction function $P$ and the parameters $\alpha$, $\beta$, and $\delta$. Without a deep understanding of the target environment, blindly applying the strategy may lead to suboptimal outcomes. However, the design of this strategy strikes a balance between testing efficiency and resource conservation for LLMs, making it a practical approach to achieving this trade-off. While we acknowledge the current limitations of LLM-driven methods in fine-grained generation, we anticipate that future advancements in LLM intelligence will mitigate these challenges. Beyond improvements in LLMs, future research could focus on developing more effective potential prediction functions and optimizing threshold selection techniques.

\section{Conclusion} \label{sec:conclusion}
In this work, inspired by the potential of large language models, we introduce a novel LLM-driven online testing framework to generate critical testing scenarios with environmental-specific search. Our framework provides an efficient pipeline for testing decision-making policies, incorporating LLM-driven scenario generation through prompt engineering. To address the unique limitations of LLMs, we propose a multi-scale generation strategy, which significantly improves testing efficiency.

Our extensive experimental results demonstrate that the proposed method outperforms baseline approaches by uncovering a greater number of failure cases while maintaining diversity in the generated scenarios. Additionally, ablation studies confirm that the multi-scale generation strategy not only boosts testing capabilities but also reduces resource consumption. Finally, the comparison of different LLMs highlights the developmental potential and robustness of our framework, emphasizing its adaptability to various LLM capabilities.

\appendices
\section{} \label{appendix A}
\label{appendix}
We provide an example of typical request and response for ACAS Xu task in Fig. \ref{fig:request} and Fig. \ref{fig:response}.

\ifCLASSOPTIONcaptionsoff
  \newpage
\fi

\bibliographystyle{IEEEtran}
\bibliography{liter}
\end{document}